\documentclass[runningheads]{llncs}
\usepackage{graphicx}

\usepackage{tikz}
\usepackage{floatrow}
\usepackage{comment}
\usepackage{amsmath,amssymb} \usepackage{color}
\usepackage{xspace}
\usepackage{multirow}
\usepackage{booktabs}
\usepackage{bbm}
\usepackage{array}
\usepackage{caption}
\usepackage{subcaption}
\usepackage{colortbl}
\usepackage{nicefrac}
\usepackage{url}
\usepackage{makecell}
\usepackage[hidelinks,hyperindex,breaklinks]{hyperref}

\hypersetup{
  breaklinks=true,     colorlinks=false,   }

\definecolor{Gray}{gray}{0.95}
\definecolor{LightGray}{gray}{0.65}

\usepackage[accsupp]{axessibility}  

\newfloatcommand{capbtabbox}{table}[][\FBwidth]

\makeatletter

\newcommand*{\@rowstyle}{}

\newcommand*{\rowstyle}[1]{  \gdef\@rowstyle{#1}  \@rowstyle\ignorespaces}

\newcolumntype{=}{  >{\gdef\@rowstyle{}}}

\newcolumntype{+}{  >{\@rowstyle}}

\makeatother

\begin{document}

\newcommand{\vqa}{VQA v2\xspace}

\newcommand{\eff}{\textsc{eff}\xspace}

\newcommand{\softmax}{\mathrm{softmax}}

\newcommand{\figref}[1]{Fig.\xspace~\ref{#1}}
\newcommand{\tabref}[1]{Tab.\xspace~\ref{#1}}
\newcommand{\secref}[1]{Sec.\xspace~\ref{#1}}
\newcommand{\appref}[1]{Appendix\xspace~\ref{#1}}
\newcommand{\eqnref}[1]{Eq.\xspace~\ref{#1}}

\newcommand{\covatr}[1]{$\mathcal{C}@$#1\%}

\newcommand{\myparagraph}[1]{\noindent\textbf{#1}}

\newcommand{\codeurl}{\url{https://github.com/facebookresearch/reliable_vqa}\xspace}

\pagestyle{headings}
\mainmatter

\title{Reliable Visual Question Answering:\\
Abstain Rather Than Answer Incorrectly}

\titlerunning{Reliable Visual Question Answering}
\author{Spencer Whitehead\inst{1}$^{\star}$ \and Suzanne Petryk\inst{1,2}$^{\star}$  \and Vedaad Shakib\inst{2} \and Joseph Gonzalez\inst{2}\and Trevor Darrell\inst{2} \and Anna Rohrbach\inst{2} \and Marcus Rohrbach\inst{1}}
\authorrunning{S. Whitehead et al.}
\institute{Meta AI \and UC Berkeley}
\maketitle

\renewcommand{\thefootnote}{\fnsymbol{footnote}}
\footnotetext[1]{Equal contribution}
\renewcommand{\thefootnote}{\arabic{footnote}}
\setcounter{footnote}{0}

\begin{abstract}
Machine learning has advanced dramatically, narrowing the accuracy gap to humans in multimodal tasks like visual question answering (VQA).
However, while humans can say ``\textit{I don't know}'' when they are uncertain (i.e., \emph{abstain} from answering a question), such ability has been largely neglected in multimodal research, despite the importance of this problem to the usage of VQA in real settings.
In this work, we promote a problem formulation for \emph{reliable VQA}, where we prefer abstention over providing an incorrect answer. We first enable abstention capabilities for several VQA models, and analyze both their \emph{coverage}, the portion of questions answered, and \emph{risk}, the error on that portion.
For that, we explore several abstention approaches.
We find that although the best performing models achieve over 70\% accuracy on the \vqa dataset, introducing the option to abstain by directly using a model's softmax scores limits them to answering less than 7.5\% of the questions to achieve a low risk of error (i.e., 1\%).
This motivates us to utilize a multimodal selection function to directly estimate the correctness of the predicted answers, which we show can increase the coverage by, for example, 2.3$\times$ from 6.8\% to 15.6\% at 1\% risk.
While it is important to analyze both coverage and risk, these metrics have a trade-off which makes comparing VQA models challenging.
To address this, we also propose an \emph{Effective Reliability} metric for VQA that places a larger cost on incorrect answers compared to abstentions.
This new problem formulation, metric, and analysis for VQA provide the groundwork for building effective and reliable VQA models that have the self-awareness to abstain if and only if they don't know the answer.\footnote{\label{fn:code}Code and models:  \codeurl}
\end{abstract}
\section{Introduction}

Visual Question Answering (VQA) is an important task and one core application of VQA is to provide a multimodal assistant, such as one that can answer questions to help with daily tasks for a user with visual impairments~\cite{antol2015vqa,gurari2018vizwiz}.
To provide such utility, users must be able to trust the output of these tools as they may be basing decisions or actions on the output~\cite{asan2020ai_trust_healthcare,gulshan2016diabetes_diagnose_trust,lutkenhoner2013reliable_diagnostic_trust,mcknight2011trust}.
While improving the accuracy of approaches may be an important factor for trusting models, models are imperfect and will inevitably produce some incorrect answers.
In many scenarios, there is a price associated with a model giving an inaccurate answer as it may mislead the user and cause them to make a mistake that could be anywhere from mildly inconvenient to very serious.
This is especially true for the example of helping users with visual impairments, since they likely do not have a method of verifying the outputs themselves.

\begin{figure}[t]
    \centering
    \includegraphics[width=0.95\textwidth]{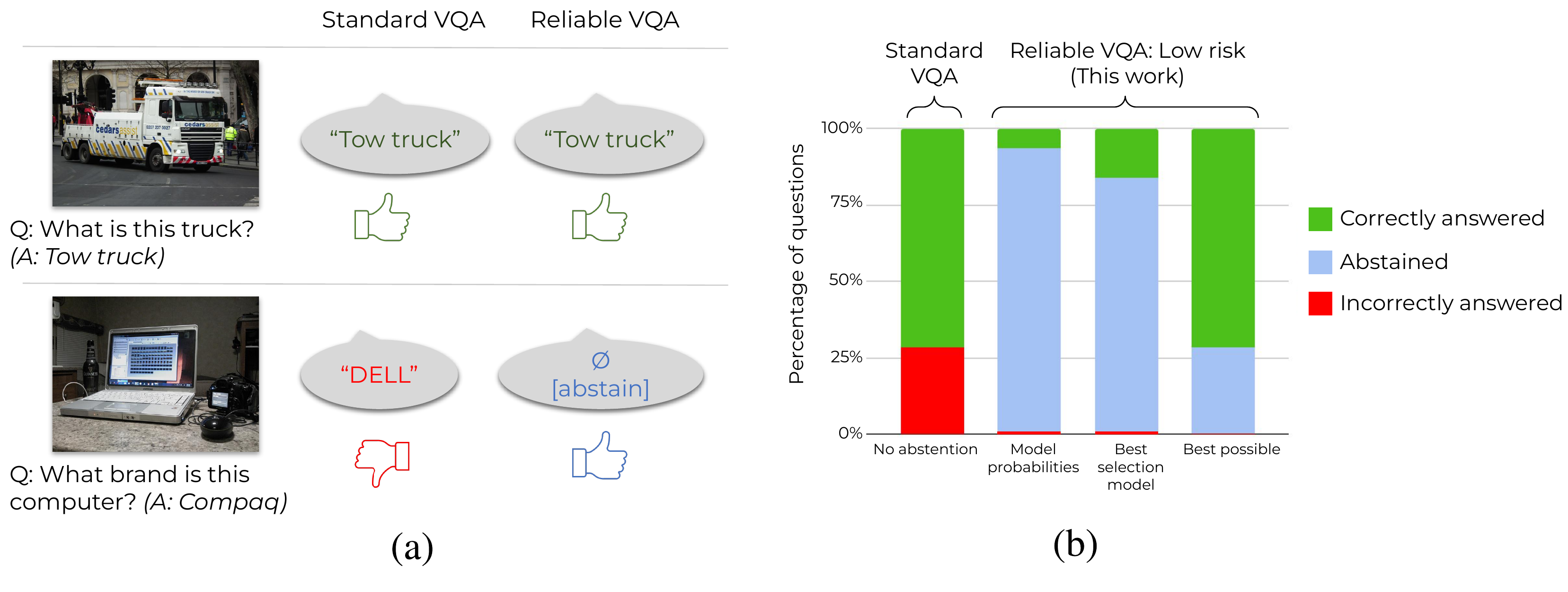}
    \caption{In the standard VQA problem, a model must answer all questions, even if it is likely to produce errors that could mislead a user, e.g., (a). A reliable VQA model, on the other hand, operates at \emph{low risk} by having the option to abstain from answering if uncertain. In (b), at 1\% risk of error, a SoTA model~\cite{shen2021much} can answer only $\sim$7\% of questions when using vanilla model probabilities to choose when to abstain. Using a learned, multimodal selection function to estimate confidences can more than double the amount of questions answered, yet there remains much room for improvement (best possible, i.e., perfect abstention).
    }
    \label{fig:teaser}
\end{figure}

One way to avoid providing incorrect information and misleading users is to \emph{abstain} from making a prediction, as in the framework of selective prediction~\cite{chow1957optimum,elyaniv2010foundations,geifman2017selective,geifman2019selectivenet}.
Consider~\figref{fig:teaser}(a): when a model is correct, we naturally would like it to give us an answer. However, when it is unable to do so (e.g., cannot ``read'' the brand name) or is very uncertain, in many application we may prefer if the model communicated ``\textit{I don't know}'', i.e., abstain~\cite{hanczar2008classification,khan2001classification}.
We say that VQA models are reliable, if they make highly accurate predictions when they choose to answer.
Ideally, reliable models should also abstain as little as possible to be effective.
Although reliability is often critical for the usage of VQA in real settings, this aspect has not received direct attention in the VQA literature aside from efforts to recognize difficult, unanswerable, or false premise questions~\cite{chiu2020assessing,gurari2018vizwiz,kafle2017analysis,ray2016question,teney2017graph}.
Also, past efforts on selective prediction have not focused on the multimodal setting, where both an image and a question can be valid or in-distribution when considered independently, yet challenging in tandem.

In this work, we formalize and explore the notion of reliability in VQA.
We propose to frame the task as a selective prediction problem~\cite{chow1957optimum,elyaniv2010foundations} in which models must either predict an answer or abstain from answering.
This requires two techniques that have not been widely explored for VQA models: (1) gauging uncertainty of predictions and (2) learning when to abstain.
To operationalize this framework, we measure performance with \emph{coverage} (how many questions are answered) and \emph{risk} (the error on these questions)~\cite{elyaniv2010foundations,kamath-etal-2020-selective}.
While low risk and high coverage are the goal, in practice there often is a trade-off between the two.
To provide a scalar measure that captures this trade-off and allows for clearer model comparisons, we introduce a new \emph{Effective Reliability} metric, which accounts for abstention while also introducing a cost for giving an incorrect answer.
This also provides an alternative evaluation for domains where it may be more intuitive to specify the penalty for an individual error instead of a bound on risk.

Under this framework, we first show that existing VQA approaches leave much room for improvement.
In particular, we demonstrate that, for a number of models, the common approach of using the maximum probability to determine abstention~\cite{hendrycks17baseline,kamath-etal-2020-selective} (by thresholding the softmax scores) limits the model to answering a small fraction of questions with a low risk of error (e.g., answering less than 7.5\% of questions at 1\% risk of error), despite having high standard VQA accuracy.
This inability to answer a larger number of questions at low risk indicates low utility of the existing VQA models.

To address this, we explore two other approaches: calibration and training a multimodal selection function.
We find that calibration often leads to a better risk-coverage trade-off compared to using the original model probabilities.
We improve beyond this by training a multimodal selection function that can better learn to predict if a the model's answer is correct, based on intermediate representations as well as the answer from the VQA model.
This selection function consistently improves the coverage of different VQA models across varying risks of error, particularly for low levels of risk.
However, we show that there is still room to improve the effectiveness of these models (see \figref{fig:teaser}(b)). Finally, we evaluate VQA models with our new Effective Reliability metric, and see that it correlates with risk and coverage in a meaningful way -- the user-defined cost of an error impacts the risk at which the model operates.

In summary, our contributions are: (1) we are the first to analyze and operationalize reliability for multimodal VQA models; (2) we expose the issue of low coverage in VQA models when asked to operate at low risk levels; (3) we explore several methods for incorporating abstention, showing that a simple yet effective multimodal selection function outperforms other methods; (4) we propose a novel \emph{Effective Reliability} metric for this problem, establishing a new benchmark for effective and reliable VQA models.

\section{Related Work}

\myparagraph{VQA methods.}
VQA is a popular task with a plethora of methods proposed in recent years~\cite{anderson2018butd,antol2015vqa,chen2020uniter,fukui2016multimodal,gao2019dynamic,jiang2020defense,jiang2018pythia,li2019visualbert,li2020oscar,lu2019vilbert,shen2021much,yang2016stacked,yu2019deep,zhang2021vinvl}.
To the best of our knowledge, there are no VQA models with a built-in abstention mechanism (i.e., they predict an answer for every image and question pair). We discuss a few exceptions with non-standard problem statements in the following.
Our work analyzes VQA models' reliability by introducing the ability to abstain into several prominent VQA models~\cite{jiang2018pythia,li2019visualbert,lu2019vilbert,shen2021much}.

\myparagraph{Detecting intrinsic difficulty.}
Some prior work on VQA involves the categorization and detection of questions that are intrinsically difficult to answer, regardless of model ability.
For example, the VizWiz VQA dataset contains labels for questions which are unanswerable~\cite{gurari2018vizwiz} and reasons for annotation entropy, such as low image quality or question ambiguity~\cite{bhattacharya2019does}.
\cite{davis2020unanswerable} define a similar categorization of unanswerable questions in VQA.
\cite{teney2017graph} compute precision/recall based on VQA model confidences and show that these can be reflective of the ambiguities of the ground truth answers.
Other work focuses on detecting whether the question incorrectly describes the visual semantics~\cite{kafle2017analysis,li2020neural,mahendru2017promise,ray2016question}.
Identifying intrinsically difficult examples has important implications in active learning, where such examples can stifle the ability of different methods to select useful examples to train on~\cite{karamcheti2021mindyouroutliers}.
In this work, we focus on predicting uncertainty specific to a model as opposed to the intrinsic difficulty from data itself.
However, in~\secref{sec:qual}, we find that a subset of questions on which a model abstains from answering are ambiguous or unanswerable.

\myparagraph{Calibration.}
In classification settings, calibration typically refers to probabilistic calibration, where the predicted confidence for a given class should be representative of the probability of the prediction being correct~\cite{guo2017calibration,hendrycks17baseline,lakshminarayanan2017simple,niculescu2005predicting,platt1999probabilistic}.
One popular parametric method is Platt scaling~\cite{platt1999probabilistic}, in which a logistic regression model is trained on classifier outputs on the validation set to return calibrated probabilities.
In our work, we explore the effectiveness of vector scaling, a multi-class extension of Platt scaling, for improving selective prediction performance.

\myparagraph{Selective prediction.}
This refers to when models have the option to abstain from providing a prediction. It is also known as sample rejection~\cite{chow1970optimum,chow1957optimum} or selective classification~\cite{elyaniv2010foundations}.
\cite{destefano00tsmc,jiang18neurips,tran2022plex} propose various related evaluation metrics.
\cite{destefano00tsmc} assigns cost coefficients to misclassified, abstained, and correctly classified samples. Concurrently with our work, \cite{tran2022plex} defines reliability as out-of-the-box performance for large-scale pretrained models across many unimodal vision or language tasks, including selective prediction.
Other works integrate abstention in multi-stage networks or ensembles~\cite{black2022selective,corbiere2019addressing,khani2016unanimous,pudil92icpr,wang2017idk}.
\cite{kadavath2022language,xin2021art} study selective prediction and transformer uncertainty within NLP tasks.
\cite{guillory2021predicting,kamath-etal-2020-selective,varshney2022investigating} explore selective prediction performance on out-of-distribution data.
\cite{kamath-etal-2020-selective} focuses on selective prediction for text-based question answering.
However, they show that their method does not generalize to questions from the same domain which are intrinsically unanswerable, whereas this represents an important portion of difficult VQA samples.
\cite{geifman2017selective,geifman2019selectivenet} optimize selective models for specific coverage levels in image classification.
We explore learned selection functions, but in the multimodal VQA setting, where the complex interaction between modalities must be modeled and more than one output may be considered correct to varying degrees.
In the multimodal space, \cite{hendricks2018women} addresses gender bias in image captioning, where the model can ``abstain'' by predicting gender-neutral words when it is uncertain.
With our proposed metric, the cost of error (e.g., misclassifying gender) can be user-defined and potentially be made class-specific.
\section{Visual Question Answering with Abstention}

Visual question answering is currently formulated and evaluated in the literature~\cite{antol2015vqa,goyal2017vqav2,gurari2018vizwiz,hudson2019gqa} as \emph{always} predicting an answer from the answer space, $\mathcal{A}$, annotated in the dataset.
So, a model $f:\mathcal{X}\mapsto\mathcal{A}$ predicts an answer $a\in \mathcal{A}$ for each input $x=(v,q)\in \mathcal{X}$, with image $v$ and question $q$.
This problem formulation forces the model to answer even if it is likely wrong, thus providing unreliable answers.
To address this, we propose to extend the VQA problem formulation so that a model is given the option to \emph{abstain} from answering a question (i.e., effectively saying ``\textit{I don't know}'').
Outside VQA, this formulation has also been referred to as ``\textit{classification with a reject option}''~\cite{chow1970optimum,destefano00tsmc,geifman2019selectivenet,hanczar2008classification,pudil92icpr} or ``\textit{selective prediction/classification}''~\cite{elyaniv2010foundations,geifman2017selective}.
We first discuss the problem definition in \secref{sec:reliability}, and then the metrics to evaluate this problem in \secref{sec:metrics}.

\subsection{Problem Definition}\label{sec:reliability}

We extend the standard VQA formulation to the setting where a model can either provide an answer from $\mathcal{A}$ or choose to abstain (denoted by $\emptyset$):  $h:\mathcal{X}\mapsto\mathcal{A}\cup\{\emptyset\}$.
We refer to $h$ as a \emph{selective model}.

One way to formulate and achieve this is by decomposing $h$ into two functions, $f$ and $g$, which jointly comprise a selective model~\cite{elyaniv2010foundations,geifman2017selective,geifman2019selectivenet}. $f$ denotes the VQA model that predicts answers and $g:\mathcal{X}\mapsto\{0,1\}$ is the selection function that determines whether the model answers or abstains from answering:
\begin{equation}\label{eq:select}
    h(x) = (f,g)(x) = 
    \begin{cases}
    f(x)& \text{if}\  g(x) = 1, \\
    \emptyset & \text{if}\ g(x) = 0. \\
    \end{cases}
\end{equation}
Given an input $x$, the selective model yields an output from $f$ when the selection function predicts that an answer should be given, or abstains if the selection function predicts that the model should not answer.
One straightforward way to formulate the selection function $g$ is based on a threshold $\gamma$, where the function $g':\mathcal{X}\mapsto[0,1]$ predicts a confidence in the correctness\footnote{While we define the output space of $g'$ as $[0,1]$ as is the case for the common softmax, one can similarly define an output space which covers, e.g., all real values $\mathbb{R}$.}  of the model $f(x)$~\cite{kamath-etal-2020-selective}:
\begin{equation}\label{eq:g-function}
    g(x) = 
    \begin{cases}
    1& \text{if}\  g'(x) \geq \gamma, \\
    0 & \text{if}\ g'(x) < \gamma. \\
    \end{cases}
\end{equation}
In general, a good function $g'(x)$ for abstention should yield high values when $f(x)$ is correct and low values when it is incorrect. In \secref{sec:approaches:selectivepredictor}, we will further discuss how to define $g'(x)$.

\subsection{Evaluation Metrics}
\label{sec:metrics}
To evaluate a VQA model with an ability to abstain, we consider two types of evaluation and discuss how we adapt them for VQA: first, \emph{coverage} and \emph{risk}~\cite{elyaniv2010foundations} and, second, a cost-based metric for balancing the two.

\myparagraph{Risk and Coverage.}
\emph{Coverage} is the portion of questions that the model opted to answer, while \emph{risk} is the error on that portion of questions~\cite{elyaniv2010foundations}.
Ideally, a reliable model should exhibit high coverage at low levels of risk, meaning it answers many questions with high accuracy and abstains on others. 
Concretely, coverage for dataset $\mathcal{D}$ with inputs $x_i$ and ground truth answers $y_i$ is given by: 
\begin{equation}\label{eq:og_coverage}
    \mathcal{C}(g) = \frac{1}{|\mathcal{D}|}\sum_{(x_i, y_i)\in\mathcal{D}} g(x_i) ,
\end{equation}
and risk is defined as:
\begin{equation}\label{eq:og_risk}
    \mathcal{R}(f, g) = \frac{\frac{1}{|\mathcal{D}|} \sum_{(x_i, y_i)\in\mathcal{D}} \ell(f(x_i), y_i)\cdot g(x_i)}{ \mathcal{C}(g)},
 \end{equation}
where $\ell$ is a cost function that measures the error between the predicted answer $f(x_i)$ and the corresponding ground truth answer $y_i$.
Assuming $g$ follows Eq.~\ref{eq:g-function}, if the threshold $\gamma$ decreases, coverage will increase, but risk will increase as well.
Hence, there is a risk-coverage trade-off that models can aim to optimize.

Applying this to VQA, the composite function $(f, g)$ becomes our selective VQA model, where $f$ produces an answer and $g$ decides whether to abstain.
However, the open-ended nature of the VQA task requires careful consideration for designing the risk-coverage metrics.
A given question might have multiple possible answers which could all be considered correct to varying degrees.
As a result, the error for a prediction on a given input is not necessarily binary.

When calculating risk, we must use a cost function that accurately represents this multi-class nature.
We follow \cite{antol2015vqa} to define VQA accuracy for a given model answer $f(x)$ as $Acc(f(x), y)=\min\left(\frac{\text{\# annotations that match }f(x)}{3}, 1\right)$ and average these accuracies over all 10 choose 9 subsets of human annotated answers for the input question, similar to other VQA evaluations~\cite{goyal2017vqav2,gurari2018vizwiz,singh2019towards}.
Under this, an answer is considered fully correct if it matches at least four of the human annotations, and receives partial credit for predicting an answer with one, two, or three humans in agreement.
Thus, our risk measurement becomes:
\begin{equation}\label{eq:vqa_risk}
    \mathcal{R}(f, g) = \frac{\frac{1}{|\mathcal{D}|} \sum_{(x_i, y_i)\in\mathcal{D}} (1- Acc(f(x_i), y_i))\cdot g(x_i)}{  \mathcal{C}(g)}.
\end{equation}

In practice, the level of risk in model predictions that a user is willing to tolerate depends highly on the scenario.
Therefore, we evaluate by computing coverage at a range of risk levels ($\mathcal{C}@\mathcal{R}$), such as coverage at 1\% or 10\% risk.
We can also summarize this over the distribution of risk levels by plotting coverage versus corresponding risk, and computing the area under this risk-coverage curve (AUC)~\cite{kamath-etal-2020-selective}.
Moreover, for an evaluation that controls for how the threshold $\gamma$ for $g$ is chosen, we compute the maximum coverage for each risk level, allowing for a more direct comparison of the selection function design.

\myparagraph{Effective Reliability.}
Recall the trade-off between risk and coverage: a standard VQA model may have high risk at 100\% coverage, but a reliable model may have low risk yet abstain on a large portion of questions (see~\figref{fig:teaser}(b)).
In practice, for a model to be reliable and effective, it should ideally achieve both low risk and high coverage.
To jointly measure these two desirable qualities, we define a metric which assigns a reward to questions that are answered correctly, a penalty to those answered entirely incorrectly, and zero reward to those abstained on.
We refer to this as \emph{Effective Reliability}, or $\Phi_c$ for a given penalty $c$, inspired by the ``effectiveness function'' introduced by \cite{destefano00tsmc}.

Formally, we define Effective Reliability for an input $x$ as $\Phi_c(x)$ (\eqnref{eq:effectiveness}), where $c$ is the cost for answering incorrectly, $g$ is the selection function, and $Acc$ is a measure of a model's correctness.
In this case, $Acc$ is the VQA accuracy~\cite{antol2015vqa}.
\begin{equation}\label{eq:effectiveness}
        \Phi_c(x) = 
    \begin{cases}
    Acc(x)& \text{if}\  g(x) = 1\  \text{and}\  Acc(x) > 0,\\
    -c & \text{if}\ g(x) = 1\  \text{and}\  Acc(x) = 0,\\
    0 & \text{if}\ g(x) = 0.
    \end{cases}
\end{equation}

\noindent We define the total score $\Phi_c = \frac{1}{n}\sum_x\Phi_c(x)$, a mean over all $n$ samples $x$.
This formulation assigns a reward to answers which are at least partially correct (i.e., $Acc(x) > 0$) -- an important property of the VQA accuracy, where the correctness of answers can vary based on the number of human annotators in agreement. The choice of $c$ depends on the deployment-specific cost of providing an incorrect answer. In \secref{sec:exp-effective}, we report $\Phi_c$ with cost values of 1, 10, and 100 ($\Phi_1$, $\Phi_{10}$, $\Phi_{100}$).
While \cite{destefano00tsmc} suggest setting $\Phi_c(x)<0$ for $g(x)=0$, we set $\Phi_c(x)=0$ (i.e., a score of 0 when abstaining).
This enables our formulation to have the clear upper bound for models which abstain perfectly (Lemma \ref{lemma}).
We provide a simple proof for this in \appref{sec:proof}.
It is also confirmed in our experiments in \tabref{tab:effective}.
\begin{lemma}
\label{lemma}
The Effective Reliability score is equal to the VQA Accuracy ($\Phi_c(x)= Acc(x)$) if a model abstains ($g(x)=0$) \emph{iff} it is incorrect ($Acc(x)=0$).
\end{lemma}

In our experiments, we choose a threshold $\gamma$ which optimizes $\Phi_c$ on a validation set to compute a model's Effective Reliability with the form of the selection function $g$ defined in \eqnref{eq:g-function}.
Additionally, the Effective Reliability score $\Phi_c$ can be evaluated for any model, even those which do not incorporate the option to abstain from providing a prediction (i.e., $g(x)$ is always 1).

Beyond its connection to VQA Accuracy (Lemma \ref{lemma}), Effective Reliability has several other advantages.
We show that it meaningfully correlates with risk-coverage (\tabref{tab:effective}), yet provides a single metric to compare models.
This offers simpler comparisons that can be used to rank approaches (e.g., evaluating on a challenge server).
It also provides an alternative evaluation for settings where it may be easier or more intuitive to define a cost for an incorrect answer as opposed to a target level of risk.

\section{Selection Functions}\label{sec:approaches:selectivepredictor}\label{sec:approaches:abstaining}
We investigate three promising directions to extend VQA models to abstain by exploring different options for $g'(x)$ introduced in \secref{sec:reliability}.
Additional implementation details for the selection functions can be found in \appref{sec:appendix:selectivefunctions}.

\myparagraph{MaxProb.}\label{sec:maxprob}
Without any additional training, a model can be extended to abstain by defining $g'$ as the softmax probability of the model's predicted class (i.e., maximum probability) and is thus refered to as MaxProb~\cite{hendrycks17baseline,kamath-etal-2020-selective,lakshminarayanan2017simple}.
Essentially, MaxProb trusts that if the model gives a high probability to one class, it is quite certain that the answer is correct and should be given: $g'_{\text{MaxProb}}(x)=\max(f'(x))$, where $f'(x)$ represents the answer probabilities.

\myparagraph{Calibration.}\label{sec:calibration}
Calibration techniques tune the absolute confidence values~\cite{platt1999probabilistic} to make the predicted probability for an output representative of the likelihood of that output being correct.
Selective prediction has more to do with relative confidence rankings~\cite{elyaniv2010foundations}, but, nevertheless, a poorly calibrated model might also imply poor confidence rankings~\cite{kamath-etal-2020-selective}.
Temperature scaling~\cite{guo2017calibration,platt1999probabilistic} is a popular calibration method, but it does not change the confidence rankings between examples and has no effect on the risk-coverage curve.
Thus, we do not consider it in this work, but instead use vector scaling~\cite{guo2017calibration,platt1999probabilistic} to calibrate the model logits.
We then apply MaxProb on top of these calibrated logits.
\appref{sec:calibraitonEval} has evaluations of how well the scores are calibrated.

\myparagraph{Multimodal selection function: Selector.}\label{sec:selectiveClassifer}
Vector scaling essentially trains an additional component on top of the VQA model to refine the model confidences.
We move beyond this by training a component (Selector) to predict whether the answer is correct~\cite{dong2018confidence,kamath-etal-2020-selective,platt1999probabilistic}.
Different from prior work on confidence estimation in other tasks~\cite{dong2018confidence,geifman2019selectivenet,kamath-etal-2020-selective,wang2017idk}, the multimodal nature of VQA presents unique challenges where the model must consider the interaction between the image, question, and answer.
To model this, we extract the image $v$, question $q$, multimodal $r$, and answer $f'(x)$ representations from the VQA model and input these to the Selector, which gives it access to representations of both the answer itself as well as the evidence on which the answer is based.
The Selector is a multi-layered perceptron that takes these representations as input and predicts the correctness of an answer with respect to the image-question pair.
To train this component, the simplest method may be to treat this as a binary classification problem (correct or incorrect).
However, this does not account for answers that may be partially correct, or where one answer may be more correct than another, as is the case with VQA.
Therefore, we propose to treat correctness prediction as a regression task where the target value is the VQA accuracy, allowing us to scale confidence scores with correctness.

\section{Experiments}

\subsection{Data and Models}
\label{sec:datamodels}
We experiment on the \vqa dataset~\cite{goyal2017vqav2} and require annotations for evaluation.
As annotations for the test-dev and test-std sets of \vqa are not publicly available, we use questions from the official validation split for our evaluation as is common~\cite{agrawal2018vqacp,cycleconsist2019,whitehead2021separating}.
As a reminder, under our selective prediction setup, the VQA model is the function $f$, the selection function is $g$, and the composition of the two form a selective model $h$.
We train the VQA models ($f$) on the training set of \vqa.
Meanwhile, we split the 214k examples in the \vqa validation set into three subsets: a split with 86k examples (40\%) for validating VQA models as well as training selection functions ($g$), another with 22k examples (10\%) for validating the selection functions, and a held out test split of 106k examples (50\%) that we use strictly for evaluating the full models ($h$).

We benchmark the selection functions introduced in \secref{sec:approaches:abstaining} in combination with VQA models with varying architectures and performance (test-std \vqa accuracy in parentheses): \textbf{Pythia}~\cite{jiang2018pythia} (70.24\%), an optimization of the widely used bottom-up top-down VQA model~\cite{anderson2018butd}; \textbf{ViLBERT}~\cite{lu2019vilbert} (70.92\%), a two-stream transformer, and \textbf{VisualBERT}~\cite{li2019visualbert} (71.00\%), a single-stream transformer, both of which use multimodal pretraining~\cite{singh2020pretrainingright}; \textbf{CLIP-ViL}~\cite{shen2021much} (74.17\%), which is the MoVie+MCAN~\cite{nguyen2021movie} model with a visual encoder from CLIP~\cite{radford2021learning}.

In \tabref{tab:benchmark}, \tabref{tab:effective}, and \figref{fig:rc_plots}, we report mean results over 10 random seeds for Pythia and CLIP-ViL (standard deviations in \appref{sec:standardDeviations}), while we report single runs for ViLBERT and VisualBERT using existing pretrained and fine-tuned models.
All other results are single runs from the same randomly chosen seed.
Details of data and model setups are in \appref{sec:additionalDatasetDetails} and \appref{sec:AdditionalModelDetails}.

\subsection{Benchmarking Risk and Coverage}\label{sec:riskcoverage}

\newcolumntype{a}{>{\color{LightGray}}c}
\begin{table*}[t]
\centering
    \resizebox{.85\columnwidth}{!}{    \begin{tabular}{=l@{\hskip5pt} +l@{\hskip6pt} +a@{\hskip10pt} +r@{\hskip6pt} +r@{\hskip6pt} +r@{\hskip6pt} +r@{\hskip10pt} +r}
    \toprule
    \multirow{2}{*}{Model $f$} & Selection & VQA  & \multicolumn{4}{c}{$\mathcal{C}@\mathcal{R}$  \ $\uparrow$} & \multirow{2}{*}{AUC  $\downarrow$} \\
    
    & function $g$ & Acc. $\uparrow$& $\mathcal{R}=1\%$ & $\mathcal{R}=5\%$ & $\mathcal{R}=10\%$ & $\mathcal{R}=20\%$ &\ \ \  \\
    
    \midrule
    
    \multirow{4}{*}{Pythia~\cite{jiang2018pythia}} & MaxProb &                                                                 64.63 & 5.84 & 24.03 & 39.71 & 68.63 & 14.53 \\
     & Calibration &                        64.90 & 6.22 & 24.37 & 40.68 & 71.29 & 14.15 \\
    & Selector &                    64.63 & \textbf{8.30} & \textbf{25.87} & \textbf{41.71} & \textbf{71.37} & \textbf{13.94} \\
    \rowstyle{\color{LightGray}}
    & Best Possible ($\mathcal{C}$) &                                        64.63 & 60.27 & 66.04 & 71.54 & 80.78 & 7.41 \\
    \midrule
    
    \multirow{4}{*}{ViLBERT~\cite{lu2019vilbert}} & MaxProb &                                                               67.51 & 7.49 & 28.56 & 46.67 & 77.40 & 12.37 \\
     & Calibration &                        67.45 & 8.81 & 29.42 & 47.24 & 77.53 & 12.22 \\
    & Selector &                   67.51 & \textbf{11.26} & \textbf{31.07} & \textbf{48.24} & \textbf{77.59} & \textbf{11.97} \\
    \rowstyle{\color{LightGray}}
    & Best Possible ($\mathcal{C}$) &                                        67.51 & 63.00 & 69.07 & 74.83 & 84.39 & 6.22 \\
    \midrule

    \multirow{4}{*}{VisualBERT~\cite{li2019visualbert}} & MaxProb &                                                                      68.44 & 6.85 & 30.34 & 49.22 & 79.33 & 11.78 \\
     & Calibration &                       68.27 & 9.72 & 31.67 & 49.68 & 79.28 & 11.63 \\
    & Selector &                   68.44 & \textbf{10.67} & \textbf{33.07} & \textbf{50.50} & \textbf{79.60} & \textbf{11.41} \\
    \rowstyle{\color{LightGray}}
    & Best Possible ($\mathcal{C}$) &                                        68.44 & 63.96 & 70.07 & 75.91 & 85.55 & 5.86 \\
    
    \midrule
    
    \multirow{4}{*}{CLIP-ViL~\cite{shen2021much}} & MaxProb &                                                                 70.01 & 6.83 & 34.08 & 54.00 & 82.30 & 10.81 \\
     & Calibration &                        69.97 & 12.43 & 36.02 & 54.03 & 82.54 & 10.55 \\
    & Selector &                   70.01 & \textbf{15.66} & \textbf{37.92} & \textbf{55.81} & \textbf{82.74} & \textbf{10.18} \\
    \rowstyle{\color{LightGray}}
    & Best Possible ($\mathcal{C}$) &                                         70.01 & 65.71 & 71.86 & 77.79 & 87.51 & 5.27 \\

    \bottomrule
    \end{tabular}}
\caption{Risk-coverage metrics for different selection functions. For coverage at risk ($\mathcal{C}@\mathcal{R}$) and VQA  Acc., higher is better. For AUC, lower is better. All in $\%$.}
\label{tab:benchmark}
\end{table*}

As discussed in \secref{sec:metrics}, we measure the maximum coverage for a given risk ($\mathcal{C}@\mathcal{R}$) as well as AUC for the risk-coverage curves and overall accuracy for each model.
We include the best possible performance on these metrics for each model, which would be a selective model that abstains only when the prediction is incorrect.
Results are reported on the test test.

\myparagraph{Selector outperforms other methods.}
From \tabref{tab:benchmark}, we see that adding the Selector consistently outperforms MaxProb in coverage for all risk tolerances as well as AUC.
The strongest improvements occur at lower risk tolerances (e.g., 1\% and 5\%), becoming smaller as the tolerance increases (e.g., 10\% and 20\%).
Notably, CLIP-ViL with Selector can improve \covatr{1} to 2.3$\times$ that of CLIP-ViL with MaxProb.
\figref{fig:rc_plots} illustrates how, for low risk levels, the addition of the selector maintains noticeably better risk as coverage increases compared to MaxProb.
It generally appears that the more accurate a model is overall, the more it may potentially improve in coverage at low risk tolerances when using Selector.
For instance, when adding the Selector, we observe the largest improvements in \covatr{1} and \covatr{5} with CLIP-ViL (8.83\% and 3.84\%, respectively), which also has the highest accuracy.
Meanwhile, Pythia has the lowest accuracy and exhibits the smallest improvements with the Selector at these tolerances (2.46\% and 1.84\%, respectively).
\figref{fig:rc_plots} depicts this between 0-5\% risk, where the gap between MaxProb and Selector appears to widen as we move to more accurate models (left to right).
Lastly, we observe that Calibration can improve coverage beyond MaxProb as well, but less than the Selector, especially at low risk tolerances (e.g., 1\%, 5\%), and not as consistently.
Because Calibration modifies the output logits, it also slightly changes model accuracy.

\newcommand{\figurewidth}{0.242}
\begin{figure*}[t]
\centering
    \includegraphics[width=\figurewidth\textwidth]{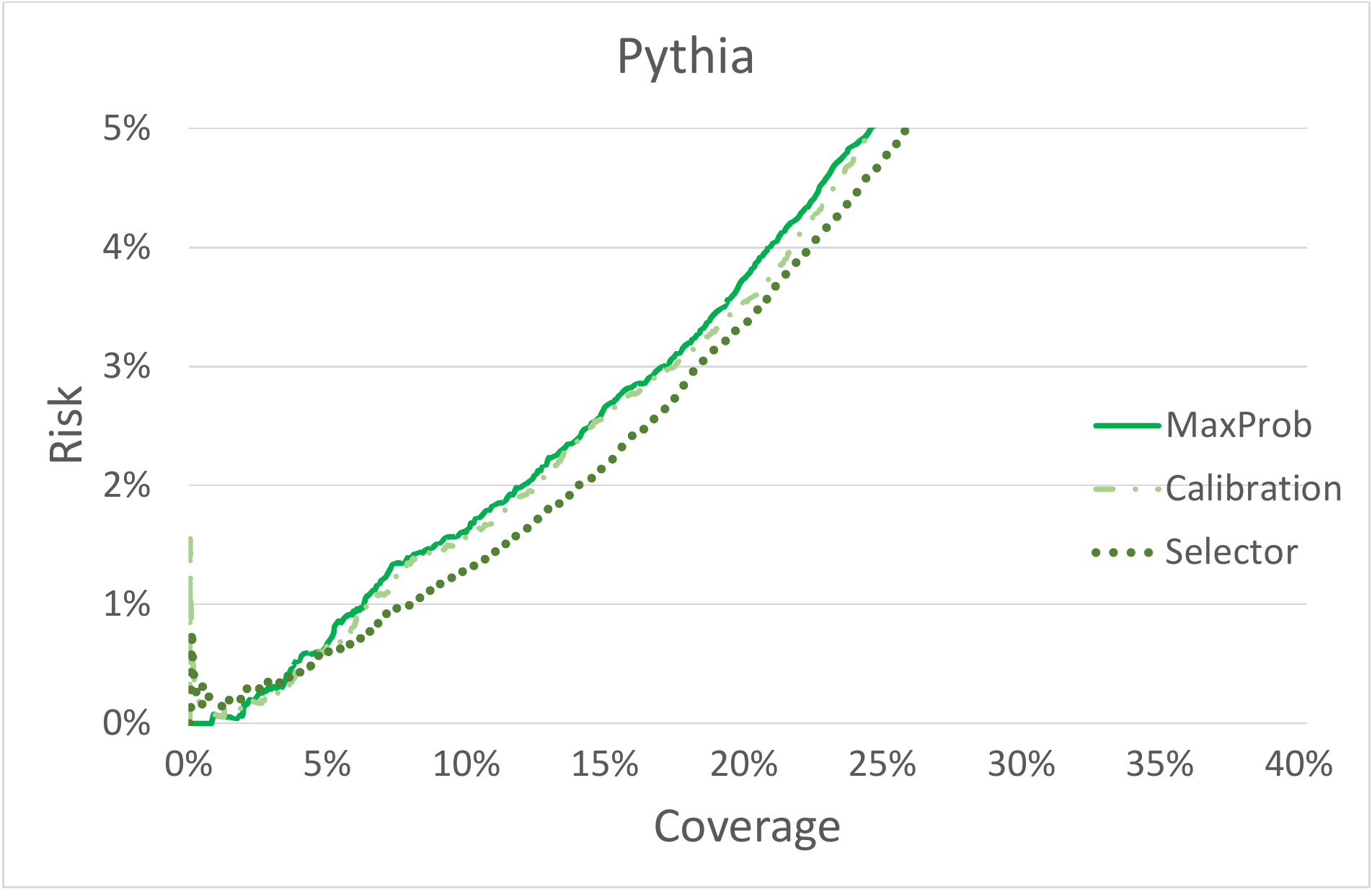}
    \includegraphics[width=\figurewidth\textwidth]{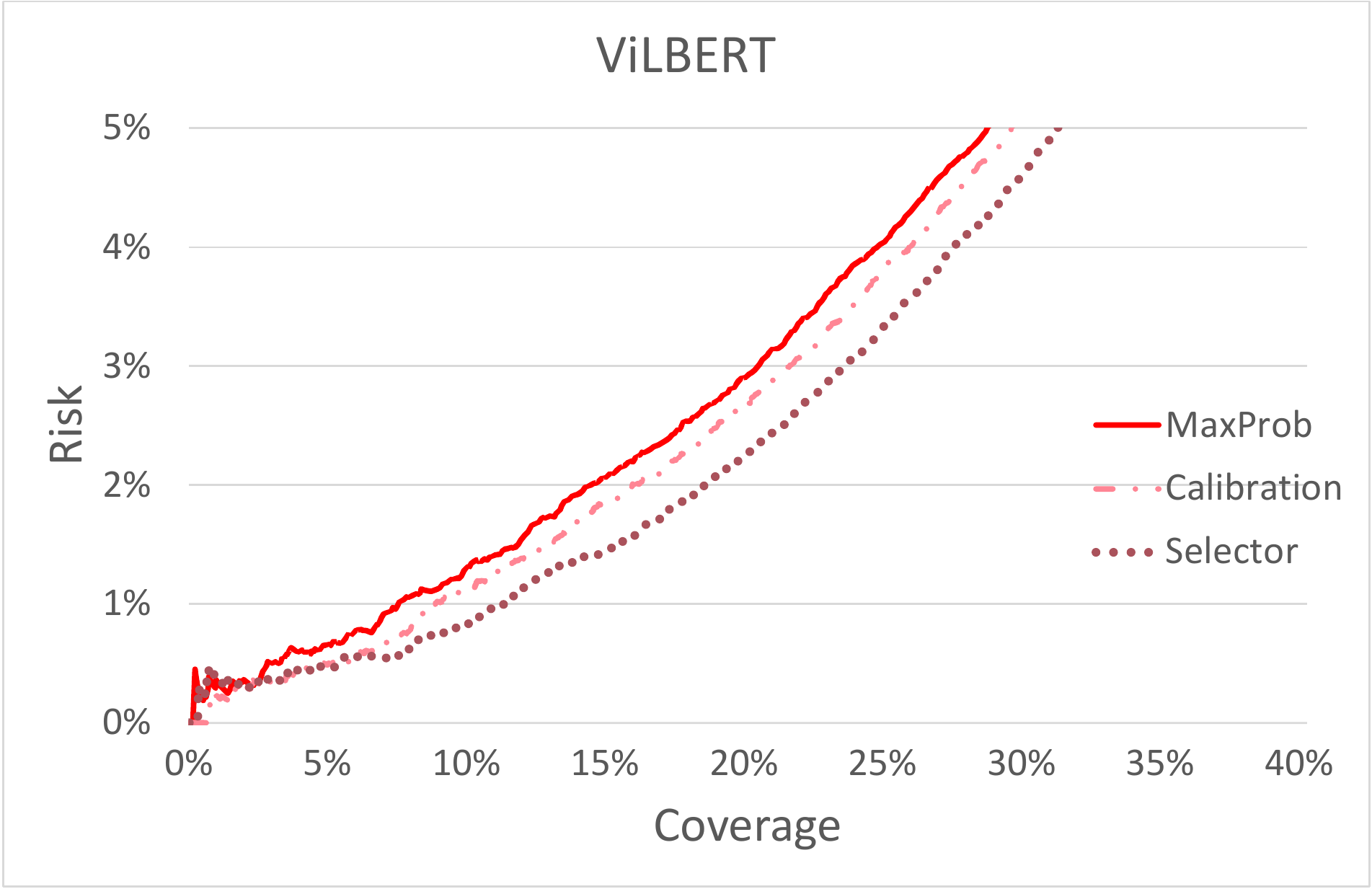}
    \includegraphics[width=\figurewidth\textwidth]{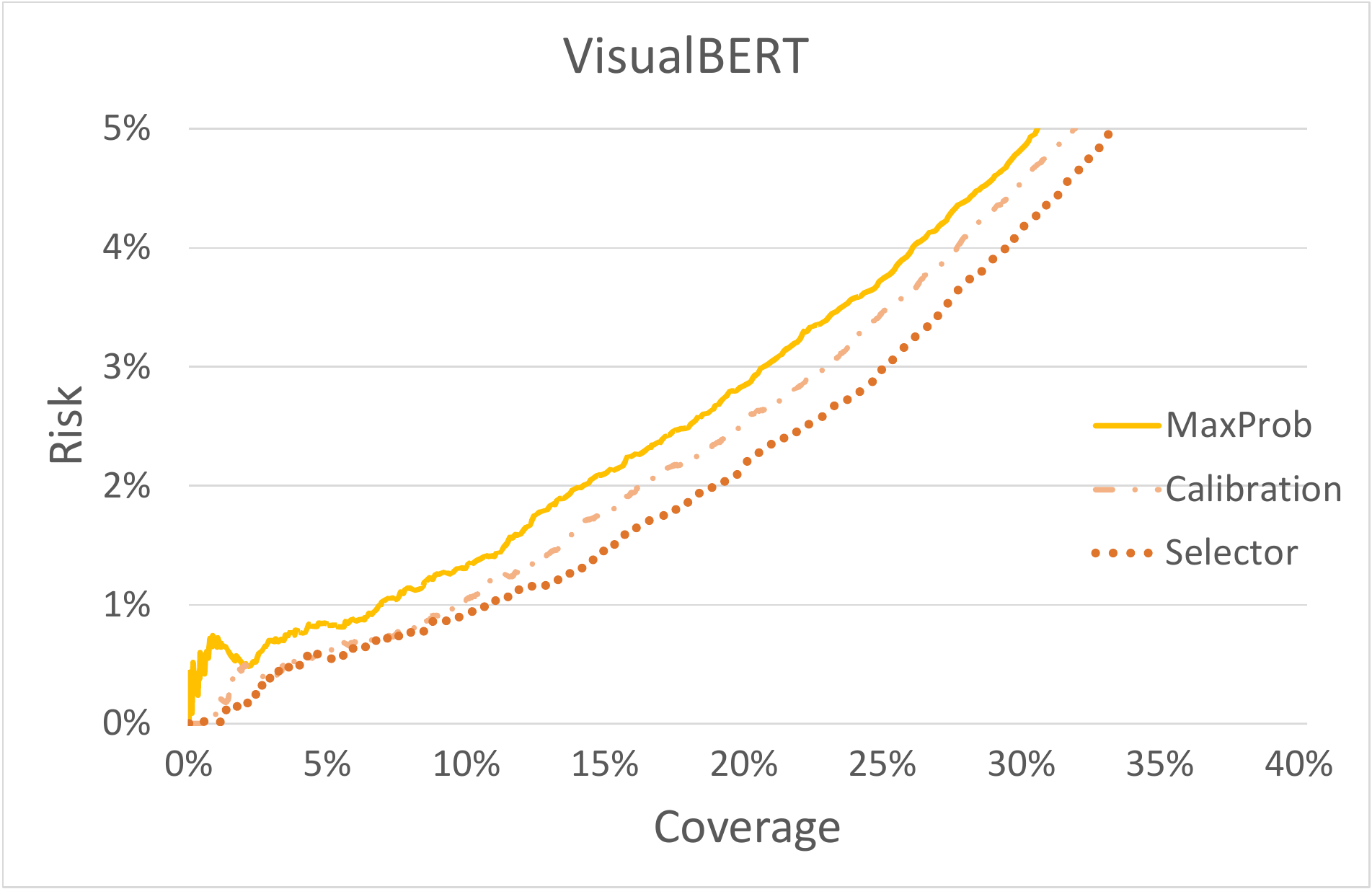}
    \includegraphics[width=\figurewidth\textwidth]{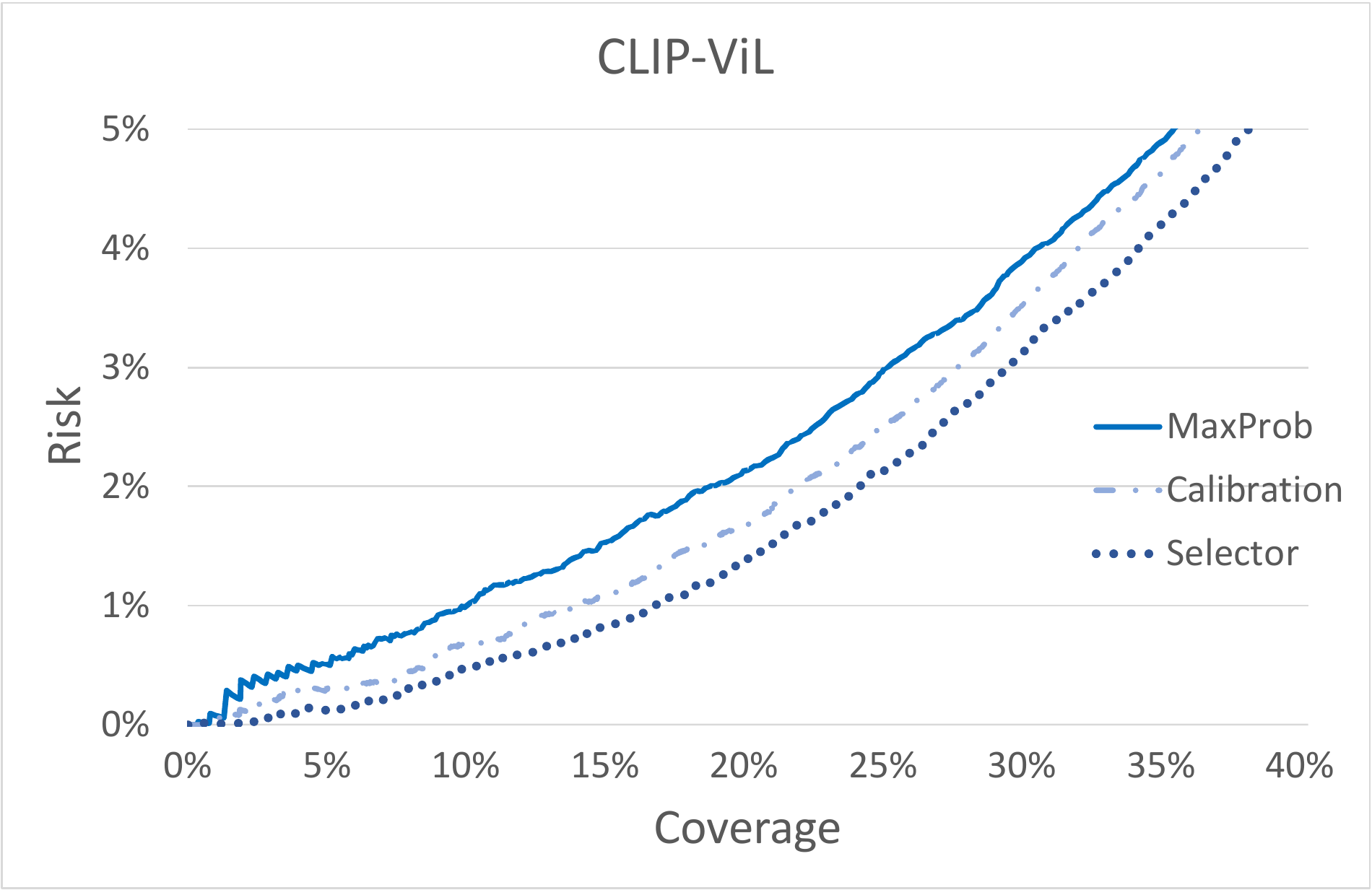}
\caption{Risk-coverage plots for each model up to 5\% risk.}
\label{fig:rc_plots}
\end{figure*}

\myparagraph{Better accuracy $\nRightarrow$ better coverage at low risk.}
While accuracy appears to positively correlate with a better risk-coverage trade-off, the results in \tabref{tab:benchmark} also imply that higher accuracy does not guarantee better coverage at low risk.
For example, CLIP-ViL has 2.50\% higher accuracy than ViLBERT, but, with default MaxProb, ViLBERT has 0.66\% higher \covatr{1} than CLIP-ViL.
\appref{sec:augmentation} also shows that augmenting the VQA model training data with the selection function training data and using MaxProb still has worse coverage at low risk than when using this data for Selector training, despite having higher accuracy.
These results imply that improving upon the risk-coverage trade-off requires not only building more accurate models but also learning better abstention policies.

\myparagraph{Still room for improvement.}
Though the evidence presented in \tabref{tab:benchmark} and \figref{fig:rc_plots} show that coverage at different risk tolerances can be improved, these approaches still fall short of the best possible.
For example, in \tabref{tab:benchmark}, the difference in \covatr{1} between each model with Selector and their respective best possibles is still $>$50\%.
Although achieving the best possible may not be realistic, more work is needed to have reliable models with high accuracy and wide coverage that shrink this gap further.

\myparagraph{Thresholds generalize to test-time.}
Thus far, we have evaluated the maximum coverage at an exact risk level.
In practice, however, a threshold $\gamma$ must be chosen, e.g., on a validation set, and used at test-time.
We evaluate how close the actual test-time risk is to the target risk when using the validation threshold with VisualBERT, with results in \appref{sec:thresholdGen}.
We find relatively small differences in risk, showing that the thresholds generalize reasonably well.
This aligns with prior findings on other tasks~\cite{geifman2019selectivenet}.
However, since the actual risks are now slightly different between models, we can no longer compare the corresponding coverages directly.
This motivates Effective Reliability, which compares models based on a predefined cost for wrong answers as opposed to an exact risk level.

\subsection{Effective Reliability}
\label{sec:exp-effective}

\newcommand{\mymidrule}{\midrule}
\begin{table*}[t]
  \centering
    \resizebox{.9\columnwidth}{!}{\begin{tabular}{=l@{\hskip5pt} +l>{\hskip6pt} +>{\columncolor{Gray}}r>{\hskip12pt} +r>{\hskip6pt} +r>{\hskip6pt} +>{\columncolor{Gray}}r>{\hskip6pt} +r>{\hskip6pt} +r>{\hskip6pt} +>{\columncolor{Gray}}r>{\hskip6pt} +r>{\hskip6pt} +r}
    \toprule
    \multirow{2}{*}{Model $f$} 
     & Selection &
    \multicolumn{3}{c}{$c$=1} & \multicolumn{3}{c}{$c$=10} & \multicolumn{3}{c}{$c$=100} \\
     & \multicolumn{1}{l}{function $g$} & \multicolumn{1}{c}{\cellcolor{Gray}$\Phi_1$ $\uparrow$}  & \multicolumn{1}{c}{$\mathcal{R}$ $\downarrow$} & \multicolumn{1}{c}{$\mathcal{C}$ $\uparrow$} & \multicolumn{1}{c}{\cellcolor{Gray}$\Phi_{10}$ $\uparrow$} & \multicolumn{1}{c}{$\mathcal{R}$ $\downarrow$} & \multicolumn{1}{c}{$\mathcal{C}$ $\uparrow$} &  \multicolumn{1}{c}{\cellcolor{Gray}$\Phi_{100}$ $\uparrow$} & \multicolumn{1}{c}{$\mathcal{R}$ $\downarrow$} & \multicolumn{1}{c}{$\mathcal{C}$ $\uparrow$} \\
    \mymidrule
    
       \multirow{4}{*}{Pythia~\cite{jiang2018pythia}} & | &                                                               36.97 & 35.37 & 100.00 & -211.96 & 35.37 & 100.00 & -2701.25 & 35.37 & 100.00 \\
    & MaxProb &                  46.49 & 22.48 & 75.58 & 15.05 & 5.68 & 26.41 & 1.90 & 0.94 & 5.13 \\ 
    & Calibration &                        47.29 & 21.66 & 74.92 & 15.18 & 5.97 & 27.73 & 2.35 & 0.92 & 5.59 \\
    & Selector &                 \textbf{47.47} & 21.02 & 73.52 & \textbf{17.03} & 6.34 & 30.16 & \textbf{3.84} & 1.01 & 8.23 \\
    \rowstyle{\color{LightGray}}
    & Best Possible ($\Phi_c$) &                                   64.63 & 10.66 & 72.34 & 64.63 & 10.66 & 72.34 & 64.63 & 10.66 & 72.34 \\ 
    
    \mymidrule
    
    \multirow{4}{*}{ViLBERT~\cite{lu2019vilbert}}  & | &                                                          42.91 & 32.49 & 100.00 & -178.51 & 32.49 & 100.00 & -2392.75 & 32.49 & 100.00 \\
    & MaxProb &                  51.50 & 21.15 & 79.92 & 17.94 & 6.45 & 34.50 & 1.67 & 1.36 & 10.18 \\      & Calibration &                      51.50 & 19.34 & 76.08 & 18.59 & 4.99 & 29.39 & 2.56 & 1.26 & 10.97 \\
    & Selector &                   \textbf{51.78} & 19.88 & 77.33 & \textbf{20.90} & 5.91 & 34.56 & \textbf{5.38} & 0.97 & 11.03 \\
    \rowstyle{\color{LightGray}}
    & Best Possible ($\Phi_c$) &                                   67.51 & 10.45 & 75.40 & 67.51 & 10.45 & 75.40 & 67.51 & 10.45 & 75.40 \\
    
    \mymidrule
    
    \multirow{4}{*}{VisualBERT~\cite{li2019visualbert}} & | &                                                                44.77 & 31.56 & 100.00 & -168.30 & 31.56 & 100.00 & -2299.01 & 31.56 & 100.00 \\
    & MaxProb &                    52.82 & 20.19 & 79.75 & 19.24 & 5.76 & 33.64 & 2.50 & 1.02 & 6.90 \\
    & Calibration &                     52.82 & 20.08 & 79.46 & 19.87 & 5.88 & 35.07 & 3.92 & 0.91 & 8.79 \\
    & Selector &                   \textbf{53.20} & 19.69 & 78.95 & \textbf{21.93} & 5.45 & 34.60 & \textbf{4.82} & 1.07 & 11.34 \\
    \rowstyle{\color{LightGray}}
    & Best Possible ($\Phi_c$) &                                   68.44 & 10.33 & 76.33 & 68.44 & 10.33 & 76.33 & 68.44 & 10.33 & 76.33 \\
    
    \mymidrule
    
    \multirow{4}{*}{CLIP-ViL~\cite{shen2021much}} & | &                                                          47.68 & 29.99 & 100.00 & -153.27 & 29.99 & 100.00 & -2162.82 & 29.99 & 100.00 \\
    & MaxProb &                  54.77 & 19.84 & 81.98 & 21.93 & 5.93 & 38.47 & 2.82 & 0.98 & 7.27 \\
    & Calibration &                        55.00 & 18.91 & 80.24 & 23.16 & 5.20 & 36.73 & 5.29 & 0.78 & 9.96 \\
    & Selector &                   \textbf{55.47} & 18.18 & 79.09 & \textbf{25.93} & 5.41 & 39.55 & \textbf{8.00} & 0.60 & 11.37 \\
    \rowstyle{\color{LightGray}}
    & Best Possible ($\Phi_c$) &                                   70.01 & 9.86 & 77.67 & 70.01 & 9.86 & 77.67 & 70.01 & 9.86 & 77.67 \\
    \bottomrule
    \end{tabular}}
\caption{Effective Reliability $\Phi_c$ for VQA models with and without abstention options. The best possible $\Phi_c$ is computed by only selecting correct predictions, and is equal to the model's VQA accuracy. All in \%.}
\label{tab:effective}
\end{table*}

We evaluate Effective Reliability ($\Phi_c$) defined in \secref{sec:reliability}, which assigns a cost to incorrect predictions, a reward to correct predictions, and zero to questions on which a model abstained from answering.
This provides a single measure to jointly consider reliability (i.e., low risk) and effectiveness (i.e., high coverage).
In \tabref{tab:effective}, we choose cost values $c$ of 1, 10, and 100, to observe how models compare when the consequences for providing an incorrect prediction become high.
Additionally, we can now directly compare to the original VQA formulation, where models do not have an option to abstain, denoted by a null selection function $g$. We also include $\Phi_c$ for the best possible $g$, where a model abstains exactly on those inputs which would result in incorrect predictions. As discussed in \secref{sec:reliability}, this is equivalent to the model accuracy. Results are reported on the test set, with an abstention threshold selected to optimize $\Phi_c$ on the validation set. We include the corresponding risk and coverage for the selected threshold.

\myparagraph{Selector still outperforms other methods.}
The Selector produces the highest Effective Reliability scores across all models and cost levels.
As the penalty for wrong answers increases, the gap between the performance of Selector and the next best model generally increases as well.
For example, the improvement of Selector over MaxProb for CLIP-ViL is 0.70\% for $\Phi_1$, yet it is 5.18\% for $\Phi_{100}$.
Further, the gap between Selector and MaxProb for $\Phi_{100}$ generally increases as the VQA model itself has higher accuracy (or best possible performance).
We observe a similar effect in \figref{fig:rc_plots}, where more accurate models have larger gaps in risk between Selector and MaxProb at a given coverage.

\myparagraph{Cost implicitly controls risk and coverage.} When the penalty for a wrong answer is high, one might expect a selective model to operate in the low-risk regime.
This is indeed reflected in \tabref{tab:effective}, where the range of risk levels for selective models at $\Phi_{100}$ ($\mathcal{R}\approx$ 0.6--1.3\%) is much lower than the range of risk at $\Phi_1$ ($\mathcal{R}\approx$ 18--22\%).
This directly translates to a similar trend in coverage, where selective models answer about 5--11\% of questions at $\Phi_{100}$, and about 74--82\% of questions at $\Phi_1$. This shows that Effective Reliability behaves intuitively around the influence of a user-selected cost on model risk and coverage.

\myparagraph{Human evaluation shows noise has little effect even with high cost values.}
For high costs (e.g., $c=100$), models are strongly penalized for producing incorrect predictions.
Given these strict penalties on errors, it becomes pertinent to ask to what degree noise in the annotations might be contributing to these penalties, though the potential impact of noise is certainly not unique to our evaluations and is a challenging problem in VQA~\cite{antol2015vqa,kafle2017visual,sharma2021survey}.
To see if our results for $\Phi_{100}$ are significantly affected by annotation noise, in \appref{sec:labelnoise}, we manually examine each sample where the model predictions were marked incorrect (and thus heavily penalized when computing $\Phi_{100}$).
We annotate cases where models may have been unfairly penalized and recompute $\Phi_{100}$ when removing this penalty.
We find that vast majority of incorrect predictions that contribute to these penalties are properly marked as incorrect.
We also see that label noise does slightly change the Effective Reliability scores at high cost, but the rankings between models and selection functions are preserved.

\myparagraph{All models without an abstention option perform poorly.} When the cost of a wrong answer is equal to the reward of getting an answer entirely correct ($c=1$), all models without a selection function $g$ underperform their selective model counterparts. As $c$ increases, this gap widens dramatically, with non-abstaining models reaching $\Phi_c$ values firmly in the negative range.
Meanwhile, all selective models reach a positive $\Phi_c$, even at high cost, illustrating the necessity of the abstention option for building models which are reliable and effective.

\subsection{Selection Function Ablations}\label{sec:selectorablations}
\begin{table}[t]
  \centering
     \resizebox{\columnwidth}{!}{    \begin{tabular}{=l@{} +c@{\hskip2pt} +c@{\hskip8pt} +r@{\hskip8pt} +r@{\hskip8pt} +r@{\hskip8pt} +r@{\hskip16pt} +r@{\hskip8pt} +r@{\hskip8pt} +r@{\hskip8pt} +r@{\hskip8pt}}
    \toprule
    
    \multirow{2}{*}{Features} & \multirow{2}{*}{Unimodal} & \multirow{2}{*}{Loss} & \multicolumn{4}{c}{$\mathcal{C}@\mathcal{R}$  \  $\uparrow$} & \multicolumn{1}{r}{\multirow{2}{*}{AUC $\downarrow$}} & \multicolumn{3}{c}{$\Phi_c$ $\uparrow$} \\
    
    &&& \multicolumn{1}{c}{$\mathcal{R}=1\%$} & \multicolumn{1}{c}{$\mathcal{R}=5\%$} & \multicolumn{1}{c}{$\mathcal{R}=10\%$} & \multicolumn{1}{c}{$\mathcal{R}=20\%$} &  & $c$=1 & $c$=10 & $c$=100\\

    \midrule
    
    $\tilde{v}$ & \checkmark & Regression &                                              0.00 & 0.00 & 0.00 & 10.18 & 24.91 & 47.10 & -0.01 & -0.85 \\
    $q$ & \checkmark & Regression &                                      0.02 & 10.78 & 33.97 & 76.33 & 14.06 & 51.81 & 10.25 & 0.94 \\
    $f'(x)$ && Regression &                              5.08 & 34.61 & 54.32 & 81.98 & 10.77 & 55.05 & 22.99 & 5.87 \\ 
    $v$ && Regression &                          11.41 & 35.34 & 51.45 & 80.57 & 11.01 & 53.75 & 23.79 & 6.55 \\
    $r$ && Regression &                          \textbf{13.26} & 32.88 & 51.26 & 80.23 & 11.11 & 53.37 & 22.17 & \textbf{7.76} \\
    \midrule[0.0001em]
    
    $f'(x)$+$\tilde{v}$ && Regression &                                          3.67 & 34.97 & 54.49 & 82.06 & 10.76 & 54.94 & 23.47 & 4.59 \\
    $f'(x)$+$q$ && Regression &                                   8.97 & 35.89 & 55.13 & 82.13 & 10.55 & 55.01 & 24.18 & 5.32 \\
    $f'(x)$+$r$ && Regression &                                  10.17 & 35.89 & \underline{55.19} & \underline{82.27} & 10.49 & \underline{55.15} & 24.19 & 5.51 \\
    $f'(x)$+$v$ && Regression &                                  12.34 & \textbf{37.26} & 55.12 & \textbf{82.40} & \underline{10.45} & \textbf{55.16} & \textbf{24.95} & 7.02 \\

    \midrule[0.0001em]
    
    $f'(x)$+$q$+$v$+$r$ && Classification &                                              6.51 & 34.87 & 55.16 & 81.58 & 10.69 & 54.69 & 23.14 & 4.36 \\
    
    $f'(x)$+$q$+$v$+$r$ && Regression &                                          \underline{12.92} & \underline{36.29} & \textbf{55.64} & \underline{82.27} & \textbf{10.43} & 55.13 & \underline{24.66} & \underline{7.31} \\
    
    \bottomrule
    \end{tabular}
}
\caption{Ablations of Selector with CLIP-ViL~\cite{shen2021much} on our selection function validation set. The overall best performance is in bold and second best is underlined. $f'(x)$, $q$, $\tilde{v}$, and $r$ are the answer, question, image, and multimodal representations, respectively. Note, $v$ is a question conditioned image representation that is not unimodal (see \appref{sec:unimodal} for details). All in \%.}
\label{tab:ablations}
\end{table}

\tabref{tab:ablations} provides ablations for the selection function design.
In the following, we distill the main observations.
Additional discussion is in \appref{sec:unimodal}.

\myparagraph{Selector requires multimodal input.}
\tabref{tab:ablations} shows the importance of using multimodal information for coverage at low risk levels.
When using each representation in isolation, we see that multimodal representations ($r$, $v$, and $f'(x)$) yield much stronger \covatr{1}, \covatr{5}, $\Phi_{10}$, and $\Phi_{100}$ than unimodal representations (image $\tilde{v}$ or question $q$).
For highly reliable models (\covatr{1}, $\Phi_{100}$), unimodal selection functions fail (coverage $\leq$0.02\%, $\Phi_{100}<1\%$), suggesting that building reliable and effective VQA models is a truly multimodal problem.
Combining all representations generally performs well, so we use this setup in all experiments.

\myparagraph{Regressing to VQA accuracy is important.}
We find that formulating the objective as a regression of the answer accuracy, rather than classifying whether the answer is correct, offers significant improvements (\tabref{tab:ablations}), especially at low risk.
This is likely because predicting the fine-grained accuracy allows the model to account for partially correct answers and learn to rank answers that are more correct higher, as opposed to classification where the distinction between partially correct answers is lost.

\myparagraph{Selector Architecture.}
\appref{sec:unimodal} presents results using different Selector architectures, where a less complex architecture can degrade performance, but a more complex one does not necessarily improve it.
Together with \tabref{tab:ablations}, we find that, rather than the network layout, the \emph{input} to the Selector and optimization target are more critical to the performance when using the Selector.

\subsection{Qualitative Analysis}\label{sec:qual}

\begin{figure}[t]
     \centering
     \includegraphics[width=\textwidth]{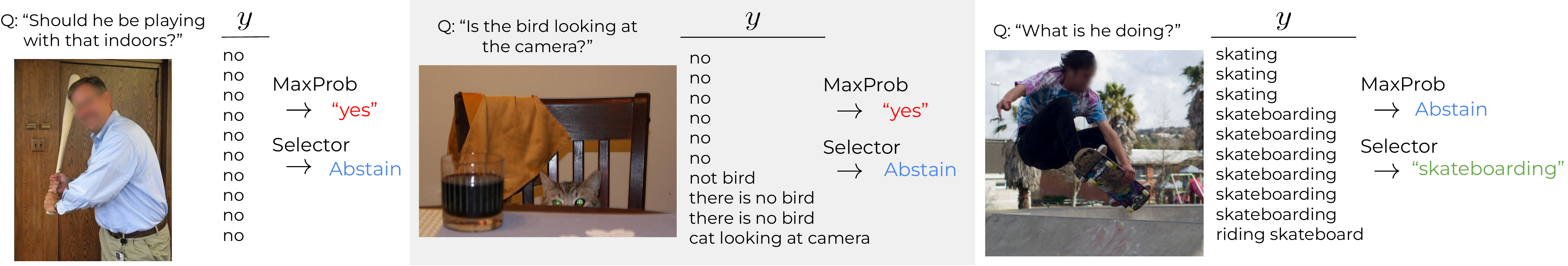}
    \caption{Qualitative test set examples with CLIP-ViL selective model predictions.}
        \label{fig:qual_analysis}   
\end{figure}

\figref{fig:qual_analysis} visualizes MaxProb and Selector decisions with CLIP-ViL for several examples on the test set (more in \appref{sec:qualitativeresults}).
The abstention threshold is chosen to maximize $\Phi_{100}$ on validation.
\figref{fig:qual_analysis} (left) shows an example of a question that requires commonsense reasoning to answer that the VQA model may not be certain of (and gets wrong), so Selector abstains.
Similarly, in \figref{fig:qual_analysis} (middle), we see a false premise question~\cite{ray2016question} where Selector abstains again as the question does not make sense for the image, while MaxProb yields an incorrect answer.
\figref{fig:qual_analysis} (right) presents an example with synonymous answers where the model is correct yet MaxProb chooses to abstain and Selector chooses to answer.
In a classification-based VQA model, synonyms can split the maximum softmax score used by MaxProb, whereas the Selector can potentially learn these answer similarities and adjust the confidence.
These examples contribute to the higher coverage at low risk observed quantitatively in our experiments.
We also find that MaxProb chooses to answer many simple questions, while Selector additionally chooses to answer more difficult, multimodal ones as well (see \appref{sec:selectordecisions}).

\section{Conclusion}
The standard VQA formulation does not include an option for models to abstain from answering if they are uncertain.
However, for many applications, it is important that the model only provides an answer if there is a low risk of error.
In this work, we promote a problem formulation for VQA which includes an option to abstain and discuss how to evaluate this, including a metric that rewards correct predictions but expects models to abstain if they are incorrect.
We benchmark several VQA models in combination with approaches for abstention.
If we want a reliable model with 1\% risk of error, we find that a state-of-the-art VQA model~\cite{shen2021much} only answers less than 7.5\% of the questions when using its softmax probabilities as estimates of model confidence.
Using calibration can improve this, but we find that the best results are consistently achieved by training a multimodal selection function to estimate correctness directly.
This increases the coverage from 6.8\% to 15.6\%.
While this is a marked improvement, one has to consider that this model achieves 70\% standard VQA accuracy on the same set of data.
With our \emph{Effective Reliability} metric, the performance drops from 70\% (for perfect abstention) to 8\% (our best abstention model) with high penalties for wrong answers.
We believe this new framework and metric for VQA will encourage the community to build VQA models which are both reliable and effective, as well as offer an opportunity for many exciting directions to improve the self-awareness of models.

{
\noindent\textbf{Acknowledgements:}
We thank Anastasios Angelopoulos and Kurt Shuster for helpful discussions. Authors, as part of their affiliation with UC Berkeley, were supported in part by the NSF CISE Expeditions Award CCF-1730628; DoD, including DARPA's LwLL, PTG, and/or SemaFor programs; the Berkeley Artificial Intelligence Research (BAIR) industrial alliance program  as well as gifts from Amazon Web Services, Ant Group, Ericsson, Facebook, Futurewei, Google, Intel, Microsoft, Scotiabank, and VMware.
}

\bibliographystyle{splncs04}
\bibliography{reliableMM}

\newpage
\clearpage
\begin{center}
{\bf {\Large Appendix to\\
Reliable Visual Question Answering:\\
Abstain Rather Than Answer Incorrectly\\} }
\end{center}
\appendix

\begin{description}
    \item[\appref{sec:changelog}] has a changelog of the versions of this paper. 
    \item[\appref{sec:unimodal}] has more discussion on  Selector ablations. 
    \item[\appref{sec:augmentation}] shows an experiment with data augmentation for MaxProb. 
    \item[\appref{sec:labelnoise}] provides a manual evaluation of the label noise.
    \item[\appref{sec:selectordecisions}] gives further analysis comparing Selector versus MaxProb decisions.
    \item[\appref{sec:qualitativeresults}] provides more qualitative results.
    \item[\appref{sec:thresholdGen}] presents results on threshold generalization.
    \item[\appref{sec:calibraitonEval}] looks at the calibration metric ECE.
    
    \item[\appref{sec:additionalDatasetDetails}] has additional details on the dataset splits.
    \item[\appref{sec:AdditionalModelDetails}] has additional model details.

    \item[\appref{sec:standardDeviations}] provides standard deviations for results in \tabref{tab:benchmark} and \tabref{tab:effective}.
    \item[\appref{sec:proof}] provides a proof of Lemma 1, providing a motivation for the definition of the Effective Reliability score $\Phi_c$.
    \item[\appref{sec:conformalprediction}] discusses the relevance of related conformal prediction works.

\end{description}

\section{Changelog}
\label{sec:changelog}
\begin{description}
\item[arXiv:2204.13631v1] First version.
\item[arXiv:2204.13631v2] Camera ready for ECCV 2022: link to publicly available code, mean and standard deviations over 10 runs for all Pythia and CLIP-ViL models, further ablations on augmentation and Selector architecture, further analysis of Selector versus MaxProb decisions, and other presentation improvements.
\item[arXiv:2204.13631v3] Previous versions unintentionally used the default ground truth answers in MMF~\cite{singh2020mmf} in the field ``\textit{answers}'' which has slightly different reference answers (e.g., it replaces some entries if they are not in the vocabulary of 3k answers). This version corrects the evaluation to use the original \vqa dataset annotations as references for evaluation.
\end{description}

\section{Selector Design Ablations}
\label{sec:unimodal}

Extending the discussion in \secref{sec:selectorablations}, we are isolating the effects of different features/modalities on the risk-coverage trade-off when using Selector.
In this direction, we experiment with different input representation variants from CLIP-ViL~\cite{shen2021much} in \tabref{tab:ablations} by ablating the question $q$, multimodal $r$, and answer $f'(x)$ representations as well as different image representations.
For image representations, we ablate the usage of the visual representation $\tilde{v}$ directly from the CLIP visual encoder~\cite{radford2021learning}, as well as the visual representation $v$ that is the concatenation of the respective pooled outputs from MCAN's self-guided attention module~\cite{yu2019deep} and MoVie's modulated convolutional bottleneck~\cite{nguyen2021movie}, which are visual representations that also contain multimodal information from the question.
Question representations are taken from the output of MCAN's self-attention module.
The multimodal representation is the concatentation of the multimodal representations that are used as inputs to the softmax output (i.e., classification) layer of CLIP-ViL.
For the answer representation, we use the logits just before the softmax in the output layer.

The results in \tabref{tab:ablations} show the importance of using multimodal information for coverage at low risk levels.
When comparing using each representation in isolation, we see that multimodal representations ($r$, $v$, and $f'(x)$) yield much stronger \covatr{1}, \covatr{5}, $\Phi_{10}$ and $\Phi_{100}$ than unimodal representations ($\tilde{v}$ and $q$).
We also observe that the answer representation achieves the best performance for \covatr{10} and \covatr{20} when each input representation is used in isolation.
Overall, we find that considering multimodal information (i.e., combinations of multimodal representations and unimodal representations from different modalities) to be most effective, with the top performers being the models that incorporate the answer representation alongside multimodal representations ($f'(x)$+$r$, $f'(x)$+$v$, and $f'(x)$+$q$+$v$+$r$).

Lastly, we also experiment with other architectures for the Selector using the same features as above.
Our Selector is a 2-layer multi-layered perceptron (MLP) (\appref{sec:appendix:selectivefunctions}).
In \tabref{tab:selectorarchitecture}, we see that a simpler, 1-layer Selector has slightly higher $\Phi_{100}$, yet lowers \covatr{1} by about 2.5\%. A more complex Transformer yields comparable performance to our 2-layer Selector.
Given these results as well as those in \tabref{tab:ablations}, we observe that the input representations and training objectives appear to be most important, and efforts for improving learned selection function performance can potentially focus on these.

\begin{table}[t]
    \centering
    \begin{tabular}{l@{\hskip8pt} c@{\hskip8pt} c@{\hskip8pt} c}
    \multicolumn{1}{c}{Architecture} & \covatr{1} $\uparrow$ & AUC $\downarrow$ &  $\Phi_{100}$ $\uparrow$  \\

    \hline
                
    1-layer Linear & 10.38 & 11.32 & \textbf{7.47}  \\
    2-layer MLP (ours) & 12.92 & 10.43 & 7.31  \\
    4-layer Transformer & \textbf{13.25} & \textbf{10.41} & 7.34 \\
    
    \end{tabular}
    \caption{Different Selector architectures with CLIP-ViL on our selection function validation split (Val in \tabref{tab:datasplits}). All in \%.}
    \label{tab:selectorarchitecture}
\end{table}

\section{Comparing to Data Augmentation}\label{sec:augmentation}

In our experiments, we use a separate set to validate VQA models and train the selection functions (Dev in \tabref{tab:datasplits}).
However, one could use this data to augment the VQA training data, which could potentially improve performance for MaxProb as there is a relationship between accuracy and these reliability metrics (\secref{sec:riskcoverage}).
\tabref{tab:augmentation} presents these results where we see that using this data to train the Selector is more effective for improving coverage at low risk levels and $\Phi_c$ with a high cost.
Since the extra data helps improve accuracy, as the risk tolerance nears the error rate of the model and coverage approaches 100\%, MaxProb surpasses Selector in coverage (i.e., \covatr{20}) and Effective Reliability (i.e., $\Phi_1$).
However, overall, these results suggest that using this data to train a Selector can be more beneficial to model reliability than using it for augmentation.

\begin{table}[t]
  \centering
\resizebox{.95\columnwidth}{!}{
\begin{tabular}{=l@{\hskip5pt} +l@{\hskip6pt} +a@{\hskip10pt} +r@{\hskip6pt} +r@{\hskip6pt} +r@{\hskip6pt} +r@{\hskip10pt} +r@{\hskip6pt} +r@{\hskip6pt} +r@{\hskip6pt} +r@{\hskip6pt}}
    \toprule
    \multirow{2}{*}{Model $f$} & Selection & \multirow{2}{*}{Acc. $\uparrow$} & \multicolumn{4}{c}{$\mathcal{C}@\mathcal{R}$  \ $\uparrow$} & \multirow{2}{*}{AUC  $\downarrow$} &  \multicolumn{3}{c}{$\Phi_c$ \ $\uparrow$} \\
    
    & \multicolumn{1}{l}{function $g$} & & $\mathcal{R}=1\%$ & $\mathcal{R}=5\%$ & $\mathcal{R}=10\%$ & $\mathcal{R}=20\%$ &  & $c$=1 & $c$=10 & $c$=100 \\
    
    \midrule
    
        \multirow{3}{*}{CLIP-ViL} & MaxProb & 69.70 & 3.32 & 31.30 & 52.57 & 81.21 & 11.13 & 54.05 & 20.17 & 1.60 \\
    
        & MaxProb-Aug & 70.52 & 6.57 & 33.01 & 54.72 & \textbf{83.25} & 10.73 & \textbf{55.61} & 22.05 & 2.76 \\
    
        & Selector & 69.70 & \textbf{12.92} & \textbf{36.29} & \textbf{55.64} & 82.27 & \textbf{10.43} & 55.13 & \textbf{24.66} & \textbf{7.31} \\

    \bottomrule
    \end{tabular}
    
}
\caption{Comparison between augmenting the training data of CLIP-ViL with our Dev set for MaxProb versus utilizing it for training Selector. Results are on our selection function validation split (Val in \tabref{tab:datasplits}). All in \%.}
\label{tab:augmentation}
\end{table}

\begin{figure}[t]
     \centering
     \includegraphics[width=\textwidth]{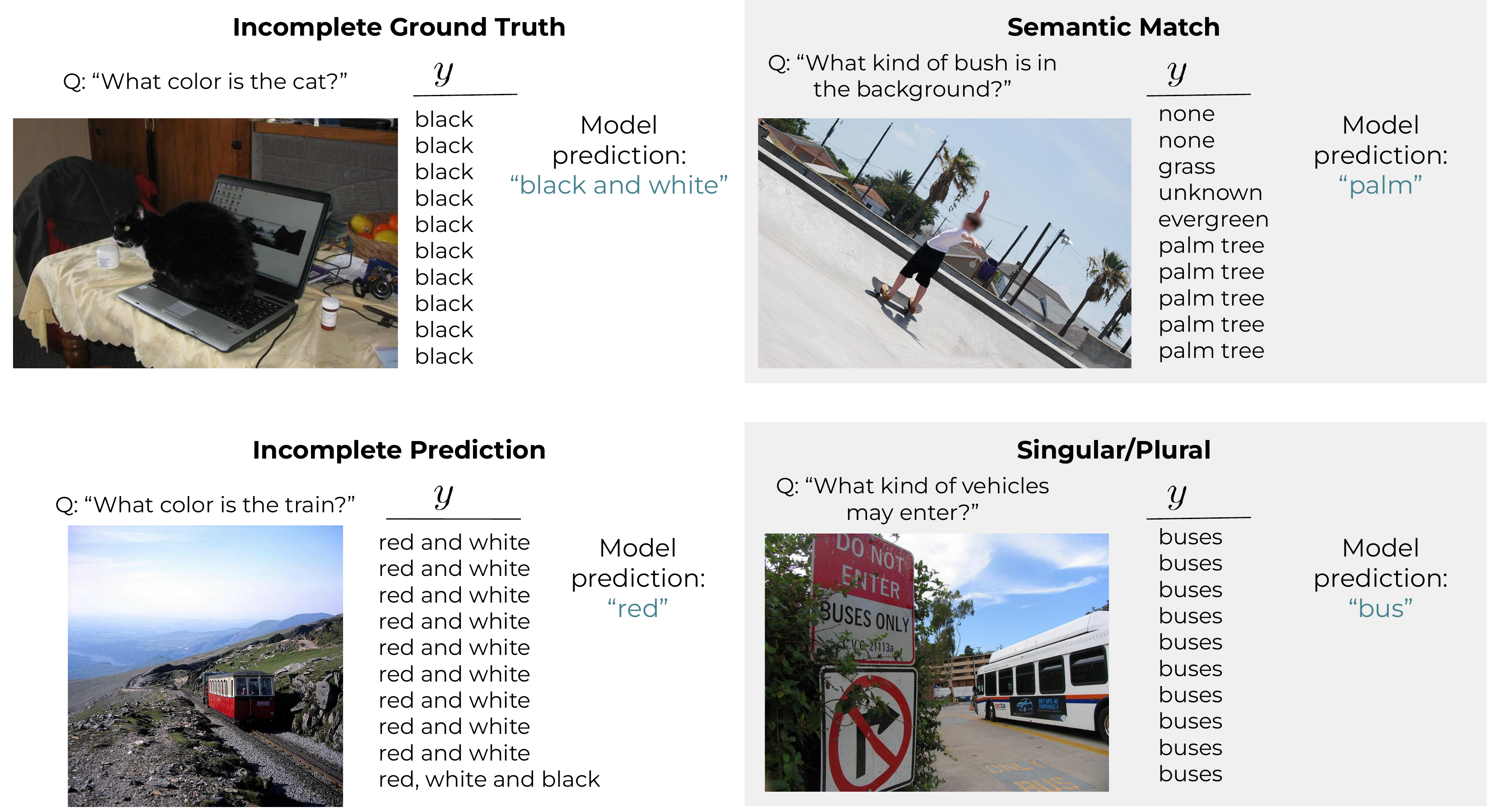}
        \caption{Example questions, images, annotations, and model predictions for each category of label noise we discover.}
        \label{fig:supplemental_labelnoise}   
\end{figure}

\section{Manual Evaluation of Label Noise}
\label{sec:labelnoise}
As discussed in \secref{sec:exp-effective}, we provide further details on our manual annotation for label noise as well as $\Phi_{100}$ when accounting for cases where the model may have been unfairly penalized. 
We specifically annotate image-question-answer triples, and discovered the following cases (\figref{fig:supplemental_labelnoise} provides examples of each):

\myparagraph{Incomplete Ground Truth}: The ground truth is in some way incomplete and simply misses the predicted answer.

\myparagraph{Semantic Match}: The predicted answer is semantically correct but does not exactly match the ground truth.

\myparagraph{Incomplete Prediction}: The predicted answer is incomplete but has part of the correct answer.

\myparagraph{Singular/Plural}: The predicted answer is singular/plural while the ground truth is plural/singular (though only if providing the opposite singular/plural version is still correct).

We do these annotations for each considered VQA model and selection function trained to optimize $\Phi_{100}$ (i.e., the strongest penalty for wrong answers) and focus our efforts on questions with VQA accuracy of 0, meaning questions that contribute negatively to $\Phi_{100}$. Once we have the annotations of unfairly penalized questions, we recompute the Effective Reliability score  $\Phi_{100}'$ when counting those questions as either abstentions or as answered questions that achieved a VQA accuracy of 100\%. Although the selection function decided to answer each of the unfairly penalized questions that we annotated, we compute $\Phi_{100}'$ under these two cases because it is unclear exactly how correct these non-matching answers should be considered. Counting them as abstentions serves as a lower bound for $\Phi_{100}'$, whereas assigning a VQA accuracy of 100\% is an upper bound.

We present the results before ($\Phi_{100}$) and after ($\Phi_{100}'$) controlling for noise in \tabref{tab:labelnoise}.
We find that while this noise does contribute to some differences in performance, it does not affect the rankings between selection functions.
For example, relative to each $\Phi_{100}$ with CLIP-ViL,  and counting unfairly penalized questions as abstentions, $\Phi_{100}'$ yields an increase of 0.37\% for MaxProb, 0.47\% for Calibration, and 0.57\% for Selector, yet the rankings remain the same.
Qualitatively, we observe that there tends to be a very significant overlap in unfairly penalized examples between selection functions, which is likely part of why the rankings remain the same.
Moreover, the amount of these label errors tends to be small, and the vast majority of questions contributing to the penalties in $\Phi_{100}$ across all models are properly marked as incorrect ($\sim$93\%).
Since the score for an incorrect sample (-100) is considerably lower than a sample marked as 100\% correct (+1), there is also little difference in $\Phi_{100}'$ when considering these few unfairly penalized questions as abstentions versus as correct answers.
These results imply that the comparisons between different selection functions at high cost (or low risk) for a given model are still meaningful despite the potential presence of noise.

\begin{table}[t]
  \centering
  \begin{tabular}{l@{\hskip6pt}l@{\hskip6pt}r@{\hskip12pt}r@{\hskip6pt}r@{\hskip12pt}r@{\hskip12pt}r@{\hskip12pt}r@{\hskip12pt}r@{\hskip12pt}r}
  \toprule
    \multirow{2}{*}{Model $f$} &  \multicolumn{1}{l}{Selection} & \multicolumn{1}{c}{\multirow{2}{*}{\% Correct GT}} & \multicolumn{1}{c}{\multirow{2}{*}{$\Phi_{100}$ $\uparrow$}} & \multicolumn{2}{c}{$\Phi_{100}'$ $\uparrow$} \\
     & \multicolumn{1}{l}{function $g$} & & & \multicolumn{1}{c}{\emph{Abstain}} & \multicolumn{1}{c}{\emph{Correct}}  \\
     \mymidrule
 \multirow{3}{*}{Pythia~\cite{jiang2018pythia}} & MaxProb & 91.30 &  1.76 & 1.95  & 1.95  \\
     & Calibration & 93.55 & 2.19  & 2.37  &  2.38 \\
     & Selector & 87.50 & \textbf{4.11}  & \textbf{4.48} & \textbf{4.49}  \\
      \mymidrule
    \multirow{3}{*}{ViLBERT~\cite{lu2019vilbert}} & MaxProb & 97.75 & 1.67  & 1.86  & 1.86 \\
     & Calibration & 94.94 & 2.56 & 2.93  &  2.94 \\
     &  Selector & 88.14 & \textbf{5.38} & \textbf{6.32} & \textbf{6.33}  \\
      \mymidrule
    \multirow{3}{*}{VisualBERT~\cite{li2019visualbert}} & MaxProb & 100.00 & 2.50  & 2.50  & 2.50  \\
     & Calibration &  97.92 & 3.92  & 4.01  & 4.01  \\
     &  Selector & 85.29 & \textbf{4.82} &\textbf{5.29} & \textbf{5.30}  \\
      \mymidrule
    \multirow{3}{*}{CLIP-ViL~\cite{shen2021much}} & MaxProb & 94.74 & 1.32 & 1.69  & 1.70 \\
     & Calibration &  93.44 & 5.32 & 5.79 & 5.80   \\
     &  Selector & 87.23 & \textbf{8.74} & \textbf{9.31} & \textbf{9.31}  \\
    \bottomrule
  \end{tabular}
  \caption{Effect of label noise on $\Phi_{100}$. \% Correct GT indicates the percentage of answered samples with a VQA accuracy of 0, where the ground truth and resulting VQA accuracy was considered correct based on the question, image, annotations, and model prediction.
  $\Phi_{100}$ indicates the original score, whereas $\Phi_{100}'$ indicates the score when counting answered questions where label errors led to a VQA accuracy of 0 as abstentions (\emph{Abstain}) or having a VQA accuracy of 100\% (\emph{Correct}) instead of being counted as incorrect. Although there is a small amount of label noise, it does not affect the ranking between selection functions with respect to Effective Reliability. All in \%.}
  \label{tab:labelnoise}
\end{table}

\section{Analysis of Selector Decisions}
\label{sec:selectordecisions}
\begin{figure}[t]
     \centering
     \includegraphics[width=0.6\textwidth]{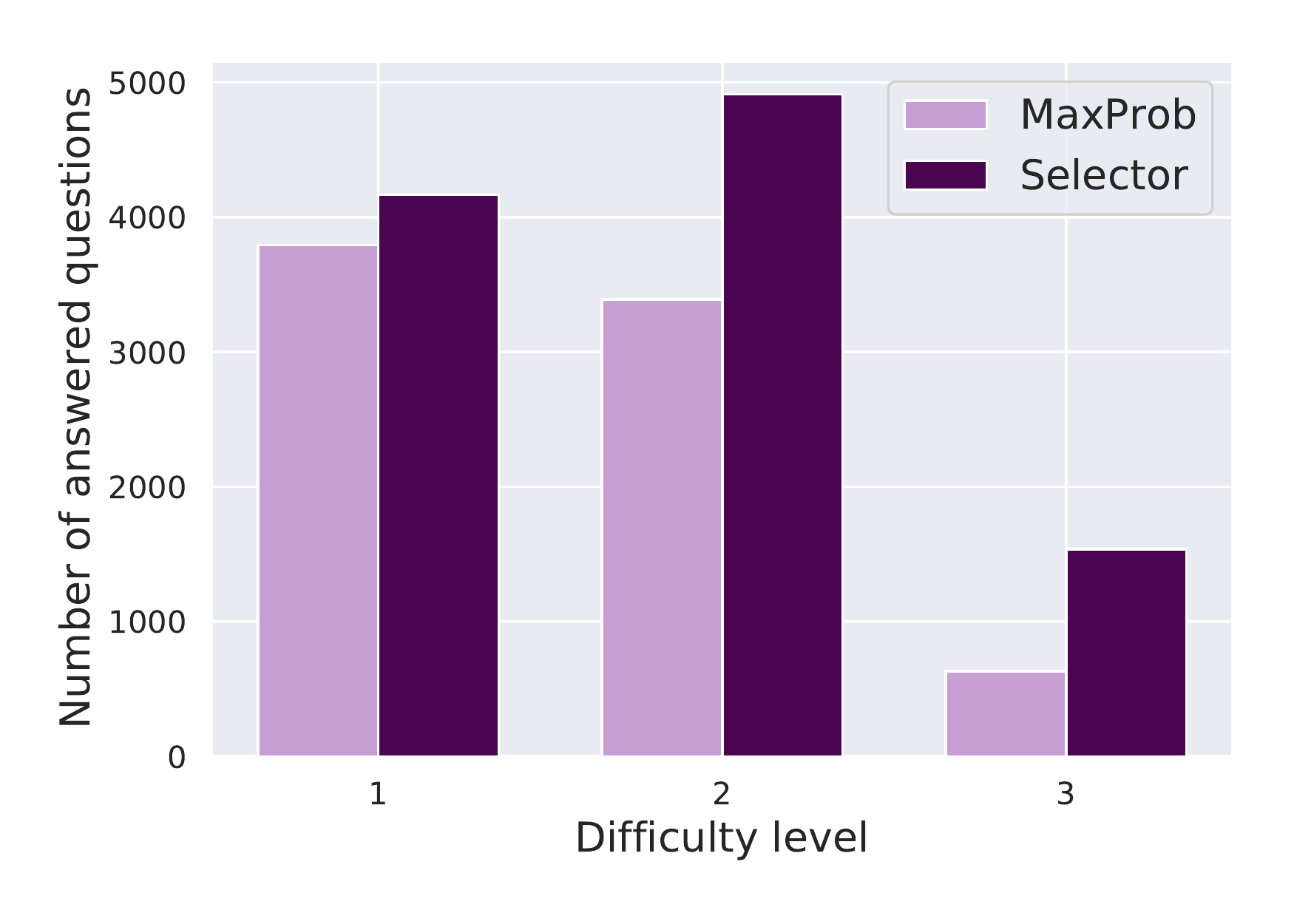}
        \caption{Number of questions in our test split that the MaxProb and Selector selection functions chose to answer, grouped by difficulty level~\cite{terao2020visual}. Level 1 corresponds to simple questions that could be answered without the image, Level 2 questions are simple to answer when considering both the question and image, and Level 3 questions are difficult to answer even when considering both modalities. Thresholds for the selection functions are chosen on the validation set to maximize $\Phi_{100}$.}
        \label{fig:selectordecisions}   
\end{figure}

We would like to understand any differences in the types of questions that the Selector chooses to abstain or answer as compared to MaxProb. We compare decisions on our test split for the two selective models, where thresholds were chosen to optimize $\Phi_{100}$ on validation. We use labels from~\cite{terao2020visual} which assign one of the following categories to each question, in order of difficulty: unimodal (Level 1), where the question could be answered without looking at the image, ``simple-multimodal'' (Level 2), where the question is simple to answer when additionally considering the image, and ``difficult-multimodal'' (Level 3), where the question is difficult to answer even when considering both modalities. \figref{fig:selectordecisions} compares the number of questions answered in each difficulty level by the MaxProb and Selector models. We find that the Selector not only answers 1.1$\times$ more unimodal questions than MaxProb, but also 1.4$\times$ more ``simple-multimodal'' and, impressively, 2.4$\times$ more ``difficult-multimodal'' questions.

\section{More Qualitative Analysis}
\label{sec:qualitativeresults}

\begin{figure}[ht!]
     \centering
     \includegraphics[width=\textwidth]{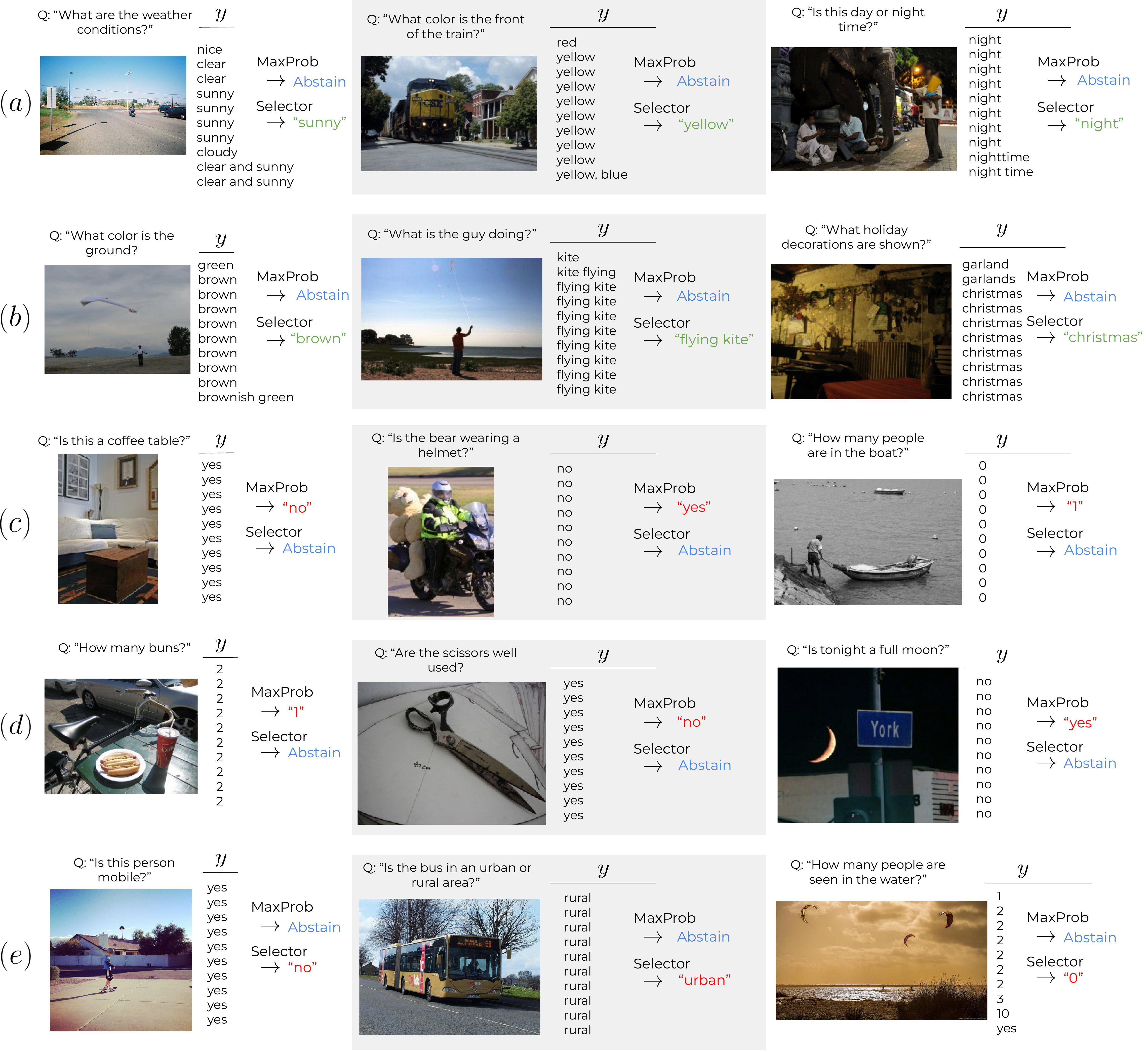}
        \caption{More qualitative test set examples with CLIP-ViL selective model predictions, when optimized for $\Phi_{100}$ on validation. Rows (a) and (b) show cases where the model was correct, yet MaxProb chose to abstain and Selector chose to answer. Rows (c) and (d) show examples of the opposite case, where the model was wrong, yet MaxProb chose to answer (contributing to the risk) and Selector chose to abstain. Row (e) shows failure cases of Selector, which chose to answer on an incorrect sample when MaxProb chose to abstain.}
        \label{fig:supplemental_qual}   
\end{figure}

In \figref{fig:supplemental_qual}, we show several more examples of cases from our test split that illustrate Selector and MaxProb decisions, where we use CLIP-ViL with selection functions optimized for $\Phi_{100}$ on the validation set (same as \figref{fig:qual_analysis}).
In particular, we show cases where the decisions of Selector and MaxProb differed --- where Selector chooses to answer while MaxProb abstains, and vice-versa.
We see some cases where the MaxProb decision to abstain may have been influenced by variability in possible answers that may cause model confidence values to be split, yet the annotations themselves have underlying semantic agreement (e.g., \figref{fig:supplemental_qual} top left, where ``\textit{sunny}'' weather conditions are also described as ``\textit{nice}'' or ``\textit{clear}'').
On the other hand, we also see cases where the model was incorrect on questions which may have been unclear or surprising, and Selector chose to abstain whereas MaxProb chose to answer (e.g., the second example on row (c) asks the unusual question ``\textit{Is the bear wearing a helmet?}'').
In these cases, we would expect a selective VQA model to abstain from answering to avoid providing an incorrect answer.
Additionally, we show several failure cases of Selector, which chose to answer on an incorrect question while MaxProb chose to abstain.

\section{Threshold Generalization}\label{sec:thresholdGen}

As discussed in \secref{sec:riskcoverage}, we evaluate how well a threshold selected for a target risk level on validation can achieve a similar level of risk on our test split. 
Experimenting with VisualBERT, comparing MaxProb and Selector, we see in \tabref{tab:risk-generalization} that the differences in risk for both selection functions tend to be at most 0.26\%. Likewise, we observe corresponding differences in achieved coverage between the validation threshold and the maximum coverage ($\Delta\mathcal{C}$). This demonstrates that the thresholds can generalize reasonably well, although it does not allow for a direct comparison of coverage for the same risk.
Effective Reliability, on the other hand, can use thresholds chosen from validation and still result in a clear comparison of models as it is a single metric.

\begin{table*}[t]
  \centering
    \resizebox{\columnwidth}{!}{
    \begin{tabular}{l@{\hskip12pt}r@{\hskip4pt}r@{\hskip6pt}r@{\hskip6pt}r@{\hskip6pt}r@{\hskip14pt}r@{\hskip6pt}r@{\hskip6pt}r@{\hskip6pt}r @{\hskip6pt}r@{\hskip6pt}r@{\hskip6pt}r@{\hskip6pt}r}
    \toprule
    
    Selection & & \multicolumn{4}{c}{$\Delta\mathcal{R}$ } & \multicolumn{4}{c}{$\Delta\mathcal{C}$ } \\

     \multicolumn{1}{l}{function $g$} & $\mathcal{R}=$ & 1\% & 5\% & 10\% & 20\% & 1\% & 5\% & 10\% & 20\%\\
    
    \midrule
    
             MaxProb &  & $+$0.11 & $-$0.20 & $+$0.26 & $-$0.15 & $+$0.91 & $-$0.68 & $+$0.82 & $-$0.34 \\
    Selector & & $+$0.22 & $+$0.26 & $+$0.22 & $-$0.17 & $+$2.79 & $+$0.89 & $+$0.74 & $-$0.36 \\
    \bottomrule
    \end{tabular}}
  \caption{Generalization of abstention thresholds $\gamma$ from validation to test, with VisualBERT. $\Delta\mathcal{R}$ and $\Delta\mathcal{C}$ are the differences in risk and coverage percentages, respectively, when using $\gamma$ selected for the target risk $\mathcal{R}$ on validation vs. $\gamma$ with maximum $\mathcal{C}@\mathcal{R}$.
}
  \label{tab:risk-generalization}
\end{table*}

\section{Effect of Model Calibration}
\label{sec:calibraitonEval}
We report the calibration performance of the vector scaling.
Specifically, we measure the expected calibration error (ECE)~\cite{guo2017calibration,naeini2015obtaining}, which measures the expected difference between the model confidence and accuracy.
The lower the ECE, the more that the model's confidence scores correspond to the actual accuracy of the predictions.
Note that the ECE metric is designed for single label classification problems.
To use the ECE metric for VQA, where there can be multiple possible answers for a question, we simply consider the most frequent human annotated answer as the ground truth for each question.

We see in \tabref{tab:ece} that vector scaling does indeed improve calibration for all models.
Taking this observation in combination with the improvements over MaxProb on $\mathcal{C}@\mathcal{R}$, AUC, and Effective Reliability seen in \tabref{tab:benchmark} and \tabref{tab:effective}, it appears that improving model calibration can help improve the risk-coverage trade-off.
However, as discussed in \secref{sec:calibration}, it is necessary to use calibration techniques that can change the relative confidence rankings, such as vector scaling.
\begin{table}[t]
  \centering
 \resizebox{\columnwidth}{!}{
  \begin{tabular}{l@{\hskip6pt}r@{\hskip6pt}r@{\hskip12pt}r@{\hskip6pt}r@{\hskip12pt}r@{\hskip6pt}r@{\hskip12pt}r@{\hskip6pt}r}
  \toprule
    & \multicolumn{2}{c}{Pythia} & \multicolumn{2}{c}{ViLBERT} & \multicolumn{2}{c}{VisualBERT} & \multicolumn{2}{c}{CLIP-ViL} \\
    
     & MaxProb & Calib. & MaxProb & Calib. & MaxProb & Calib. & MaxProb & Calib. \\
    \midrule
        ECE $\downarrow$ & 0.1701  & \textbf{0.0938}  & 0.1457  & \textbf{0.1121} & 0.1458 & \textbf{0.1169} & 0.1974  & \textbf{0.1522} \\
    \bottomrule
  \end{tabular}
 }
  \caption{ECE of different models with (Calibration, denoted Calib.) and without (MaxProb) the vector scaling calibration on our test split. Lower is better.}
  \label{tab:ece}
\end{table}

\section{Additional Dataset Split Details}
\label{sec:additionalDatasetDetails}

\begin{table*}[t]
  \centering
    \begin{tabular}{l@{\hskip8pt} l@{\hskip10pt} l@{\hskip8pt} r@{\hskip8pt} r@{\hskip8pt} r@{\hskip8pt} r}
    \toprule
    
    \multicolumn{1}{c}{Source} & \multicolumn{1}{c}{Split Name} & \multicolumn{1}{c}{Usage} & \% src &\multicolumn{1}{c}{\#I} & \multicolumn{1}{c}{\#Q} & \multicolumn{1}{c}{\#A} \\
    
    \midrule
    
    \vqa train & Train & Train $f$ & 100\% & 82,783 & 443,757 & 4,437,570 \\
    
    \midrule[0.0001em]
    
    \multirow{3}{*}{\vqa val} & Dev & Validate $f$ / Train $g$ & 40\% & 16,202 & 86,138 & 861,380 \\

     & Val & Validate $g$ & 10\% & 4,050 & 21,878 & 218,780 \\
     
     & Test & Test $h$ & 50\% & 20,252 & 106,338 & 1,063,380 \\

    \bottomrule
    \end{tabular}
\caption{Table of statistics for the dataset splits used for training as well as validating VQA models ($f$), training as well as validating selection functions ($g$), and testing full selective models ($h = (f, g)$). \% src indicates the percentage of the source data (Source) that each split represents. \#I, \#Q, and \#A indicate the number of images, questions, and answers, respectively.}
\label{tab:datasplits}
\end{table*}

We experiment on the \vqa dataset~\cite{goyal2017vqav2}, which contains a large amount of human-annotated image-question-answer triplets.
\tabref{tab:datasplits} lays out the data splits we use in our experiments.
We create splits of the \vqa validation set since we require answer annotations to evaluate risk, coverage, and Effective Reliability.
These splits are created such that no images (and therefore no question-answer annotations) are shared between them.
Note that the data in the held out test set (Test in \tabref{tab:datasplits}) is never seen during the training or validation of any component ($f$ or $g$) and is only used for evaluations.
All presented results are on our test set unless otherwise specified.

\section{Model Details}
\label{sec:AdditionalModelDetails}

In this section, we present the details of the models used in our experiments.

\begin{table*}[t]
  \centering
    \begin{tabular}{lr@{\hskip10pt} r@{\hskip10pt} r@{\hskip10pt} r}
    \toprule
    
    Hyperparameters & \multicolumn{1}{c}{Pythia} & \multicolumn{1}{c}{ViLBERT$^{\dagger}$} & \multicolumn{1}{c}{VisualBERT$^{\dagger}$} & \multicolumn{1}{c}{CLIP-ViL} \\
    
    \midrule
    
    Batch Size & 512 & 896 & 896 & 32 \\
    Hidden Size & 5,000 & 1,024 & 768 & 1,024 \\
    \# Layers & L-1, V-1 & L-12, V-6 & 12 & 6 / 4 \\
    Optimizer & Adamax\cite{kingma2015adam} & AdamW\cite{loshchilov2017decoupled} & AdamW\cite{loshchilov2017decoupled} & AdamW\cite{loshchilov2017decoupled} \\
    Adam $\epsilon$ & 1e-8 & 1e-8 & 1e-8 & 1e-9 \\
    Adam $\beta_1$ & 0.9 & 0.9 & 0.9 & 0.9 \\
    Adam $\beta_2$ & 0.999 & 0.98 & 0.98 & 0.98 \\
    Learning rate & 0.01 & 5e-5 & 5e-5 & 5e-5  \\
    Dropout & -- & 0.1 & 0.1 & 0.1 \\
    \# Steps & 22,000 & 88,000 & 88,000 & 236,000\\
    \# Warmup Steps & 1,000 & 2,000 & 2,000 & 54,000 \\
    Max Grad. L2-Norm & 0.25 & -- & -- & 5 \\

    \bottomrule
    \end{tabular}
\caption{Hyperparameters of each model used in our experiments. Max Grad.\ L2-Norm is used for gradient clipping. L and V indicate language and vision layers, respectively. The 6 / 4 for CLIP-ViL indicates that the model has 6 MCAN layers and 4 MoVie layers.
$\dagger$ indicates that the hyperparameters are reported directly from~\cite{singh2020pretrainingright}.
}
\label{tab:hyperparams}
\end{table*}

\subsection{VQA Models}

We use the open-source MMF framework \cite{singh2020mmf} for all our experiments, which contains implementations of each VQA model.\footnote{\url{https://mmf.sh/}}
For training VQA models, we follow the hyperparameters from MMF, which we list in \tabref{tab:hyperparams}.
All models treat VQA as a classification task and are trained with VQA accuracy as soft target scores via a binary cross-entropy loss~\cite{teney2018tips}.
We briefly discuss the models and settings used in our experiments, extending \secref{sec:datamodels}:

\myparagraph{Pythia}~\cite{jiang2018pythia}:
A previous state-of-the-art model that won the 2018 VQA challenge and is an optimization of the widely used bottom-up top-down (BUTD) VQA model~\cite{anderson2018butd}.
This model uses BUTD object detection features~\cite{anderson2018butd} trained on Visual Genome~\cite{krishna2017visual}, but the features are extracted from a ResNext-152 based FasterRCNN~\cite{ren2015faster}.
Pythia's implementation further uses grid features from a ResNet-152~\cite{he2016deep} as additional inputs to improve performance~\cite{jiang2018pythia}.
GloVe embeddings~\cite{pennington2014glove} are used to initialize the word representations.
We train this model from scratch on the \vqa training data.

\myparagraph{ViLBERT}~\cite{lu2019vilbert}:
A two-stream vision-and-language transformer model~\cite{cao2020behind,tan2019lxmert} that also uses object detection features.
The same object detection features from Pythia are used, but without the addition of grid features.
We use the pretrained and fine-tuned model provided by MMF.\footnote{\url{https://github.com/facebookresearch/mmf/tree/main/projects/pretrain_vl_right}\label{fn:mmf_ptright_url}}
The MMF version of this model is from \cite{singh2020pretrainingright} is pretrained on the \vqa training data~\cite{goyal2017vqav2} using self-supervised objectives (masked language modeling and masked image modeling).
The VQA model is initialized with the pretrained encoder weights, and then fine-tuned on the \vqa training data.

\myparagraph{VisualBERT}~\cite{li2019visualbert}:
This model is a single-stream transformer architecture, like BERT~\cite{devlin2018bert}.
Here, the setup is very similar to ViLBERT and we use the same visual features as ViLBERT.
We again use the pretrained and fine-tuned model provided by MMF.\textsuperscript{\ref{fn:mmf_ptright_url}}
This MMF version of VisualBERT~\cite{singh2020pretrainingright} is pretrained on MSCOCO captions~\cite{chen2015microsoft} using a masked language modeling objective.
Just like ViLBERT, the VQA model is also initialized with the pretrained encoder weights and fine-tuned on \vqa.

\myparagraph{CLIP-ViL}~\cite{shen2021much}:
This represents a state-of-the-art model that is trained from scratch on the VQA data whose visual encoder is from the CLIP model~\cite{radford2021learning}.
The visual representations are grid features that are obtained from the visual encoder of the CLIP model~\cite{radford2021learning}.
We use the implementation provided by the authors of \cite{shen2021much} to extract the visual features.\footnote{\url{https://github.com/clip-vil/CLIP-ViL/tree/master/CLIP-ViL-Direct/vqa}}
The VQA architecture, MoVie+MCAN~\cite{nguyen2021movie}, is an ensemble of a transformer encoder-decoder~\cite{yu2019deep} and modulated convolutional~\cite{nguyen2021movie} model, which won the 2020 VQA challenge.
GloVe embeddings~\cite{pennington2014glove} are also used to initialize the word representations.
Like Pythia, we train this VQA model from scratch on \vqa training data.

\subsection{Selection Functions}
\label{sec:appendix:selectivefunctions}

We detail the Calibration and Selector selection functions here.
We do not cover MaxProb as no additional training is required.
While training each selection function, we freeze the weights of the VQA model.

\myparagraph{Calibration.}
The inputs to the calibration are the unnormalized answer logits (i.e., answer representation just before the softmax) of the VQA model, and the outputs are the calibrated logits.
Since we use vector scaling~\cite{guo2017calibration,platt1999probabilistic}, we input the logits from the VQA model into a linear layer with a diagonal weight matrix and a bias term.
During training, after the linear layer, we apply a sigmoid activation and, in contrast to~\cite{guo2017calibration}, use these as input to a binary cross entropy loss with the soft VQA labels~\cite{teney2018tips}.
We train the linear layer using the AdamW optimizer~\cite{loshchilov2017decoupled} with a learning rate of 0.01 and a weight decay of 1e-4.
At test time, we use the output of this linear layer as our calibrated logits, apply a softmax, and use the same abstention procedure as MaxProb (\secref{sec:calibration}).

\myparagraph{Selector.}
The inputs to Selector are the answer, question, image, and multimodal representations.
For each input, we have a specific 1-layer MLP with a ReLU activation and hidden size of 512.
We then concatenate the outputs of these layers and input them to a 2-layer MLP with ReLU activations and hidden size of 1,024, followed by a binary output layer to produce a confidence value.
This architecture remains exactly the same for all models.
However, if a model produces a set of representations for the image or question, then we max pool these features to collapse them to a single representation.
For optimization, we employ the AdamW optimizer~\cite{loshchilov2017decoupled} with a learning rate of 1e-4, a batch size of 256, and gradient clipping with a max gradient L2 norm of 0.25.

\section{Extended Results}\label{sec:standardDeviations}

\tabref{tab:benchmark-appendix} and \tabref{tab:effective-appendix} provide the mean and standard deviation over the 10 random seeds for Pythia and CLIP-ViL results.
Due to difficulties reproducing the pretrained and fine-tuned performance of ViLBERT and VisualBERT, we simply use existing checkpoints in MMF\textsuperscript{\ref{fn:mmf_ptright_url}} and report single run metrics for these VQA models.

\begin{table*}[t]
\centering
    \resizebox{\columnwidth}{!}{
    \begin{tabular}{=l@{\hskip5pt} +l@{\hskip6pt} +a@{\hskip10pt} +r@{\hskip6pt} +r@{\hskip6pt} +r@{\hskip6pt} +r@{\hskip10pt} +r}
    \toprule
    \multirow{2}{*}{Model $f$} & Selection & \multirow{2}{*}{Acc. $\uparrow$} & \multicolumn{4}{c}{$\mathcal{C}@\mathcal{R}$   \ $\uparrow$} & \multirow{2}{*}{AUC  $\downarrow$} \\
    
    & \multicolumn{1}{l}{function $g$} & & $\mathcal{R}=1\%$ & $\mathcal{R}=5\%$ & $\mathcal{R}=10\%$ & $\mathcal{R}=20\%$ & \ \ \  \\
    
    \midrule
    
    \multirow{4}{*}{Pythia} & MaxProb & 
         64.63 $\pm$ 0.10 & 5.84 $\pm$ 0.36 & 24.03 $\pm$ 0.41 & 39.71 $\pm$ 0.34 & 68.63 $\pm$ 0.33 & 14.53 $\pm$ 0.08 \\
     & Calibration & 
         64.90 $\pm$ 0.09 & 6.22 $\pm$ 0.47 & 24.37 $\pm$ 0.43 & 40.68 $\pm$ 0.29 & 71.29 $\pm$ 0.25 & 14.15 $\pm$ 0.08 \\
    & Selector & 
         64.63 $\pm$ 0.10 & 8.30 $\pm$ 0.36 & 25.87 $\pm$ 0.35 & 41.71 $\pm$ 0.41 & 71.37 $\pm$ 0.22 & 13.94 $\pm$ 0.07 \\
    \rowstyle{\color{LightGray}}
    
    & Best Possible ($\mathcal{C}$) & 
         64.63 $\pm$ 0.10 & 60.27 $\pm$ 0.11 & 66.04 $\pm$ 0.12 & 71.54 $\pm$ 0.13 & 80.78 $\pm$ 0.13 & 7.41 $\pm$ 0.05 \\
    \midrule
    
    \multirow{4}{*}{CLIP-ViL} & MaxProb & 
         70.01 $\pm$ 0.13 & 6.83 $\pm$ 1.93 & 34.08 $\pm$ 1.17 & 54.00 $\pm$ 0.38 & 82.30 $\pm$ 0.20 & 10.81 $\pm$ 0.12 \\
     & Calibration & 
         69.97 $\pm$ 0.11 & 12.43 $\pm$ 0.69 & 36.02 $\pm$ 0.29 & 54.03 $\pm$ 0.37 & 82.54 $\pm$ 0.20 & 10.55 $\pm$ 0.06 \\
    & Selector & 
          70.01 $\pm$ 0.13 & 15.66 $\pm$ 0.71 & 37.92 $\pm$ 0.25 & 55.81 $\pm$ 0.41 & 82.74 $\pm$ 0.24 & 10.18 $\pm$ 0.07 \\
    \rowstyle{\color{LightGray}}
    & Best Possible ($\mathcal{C}$) & 
          70.01 $\pm$ 0.13 & 65.71 $\pm$ 0.14 & 71.86 $\pm$ 0.15 & 77.79 $\pm$ 0.14 & 87.51 $\pm$ 0.16 & 5.27 $\pm$ 0.05 \\
    \bottomrule
    \end{tabular}
    }
\caption{Mean and standard deviations for risk-coverage metrics for different selection functions from \tabref{tab:benchmark}. All in \%. See \secref{sec:standardDeviations}.}
\label{tab:benchmark-appendix}
\end{table*}

\begin{table*}[t]
  \centering
    \resizebox{1.0\columnwidth}{!}{
\begin{tabular}{=l@{\hskip5pt} +l>{\hskip6pt} +>{\columncolor{Gray}}r>{\hskip12pt} +r>{\hskip6pt} +r>{\hskip6pt} +>{\columncolor{Gray}}r>{\hskip6pt} +r>{\hskip6pt} +r>{\hskip6pt} +>{\columncolor{Gray}}r>{\hskip6pt} +r>{\hskip6pt} +r}
    \toprule
    \multirow{2}{*}{Model $f$} & Selection  &
    \multicolumn{3}{c}{$c$=1} & \multicolumn{3}{c}{$c$=10} & \multicolumn{3}{c}{$c$=100} \\
     & \multicolumn{1}{l}{function $g$} & \multicolumn{1}{c}{\cellcolor{Gray}$\Phi_1$ $\uparrow$}  & \multicolumn{1}{c}{$\mathcal{R}$ $\downarrow$} & \multicolumn{1}{c}{$\mathcal{C}$ $\uparrow$} & \multicolumn{1}{c}{\cellcolor{Gray}$\Phi_{10}$ $\uparrow$} & \multicolumn{1}{c}{$\mathcal{R}$ $\downarrow$} & \multicolumn{1}{c}{$\mathcal{C}$ $\uparrow$} &  \multicolumn{1}{c}{\cellcolor{Gray}$\Phi_{100}$ $\uparrow$} & \multicolumn{1}{c}{$\mathcal{R}$ $\downarrow$} & \multicolumn{1}{c}{$\mathcal{C}$ $\uparrow$} \\
    \mymidrule
    
       \multirow{5}{*}{Pythia} & | &                                       36.97 $\pm$ 0.19 & 35.37 $\pm$ 0.10 & 100.00 $\pm$ 0.00 & -211.96 $\pm$ 1.00 & 35.37 $\pm$ 0.10 & 100.00 $\pm$ 0.00  & -2701.25 $\pm$ 9.16 & 35.37 $\pm$ 0.10 & 100.00 $\pm$ 0.00  \\
    & MaxProb &                  46.49 $\pm$ 0.13 & 22.48 $\pm$ 0.18 & 75.58 $\pm$ 0.44 & 15.05 $\pm$ 0.34 & 5.68 $\pm$ 0.61 & 26.41 $\pm$ 1.88 & 1.90 $\pm$ 0.55 & 0.94 $\pm$ 0.31 & 5.13 $\pm$ 1.79 \\
    & Calibration &                       47.29 $\pm$ 0.15 & 21.66 $\pm$ 0.45 & 74.92 $\pm$ 0.90 & 15.18 $\pm$ 0.39 & 5.97 $\pm$ 0.77 & 27.73 $\pm$ 2.48 & 2.35 $\pm$ 0.63 & 0.92 $\pm$ 0.25 & 5.59 $\pm$ 1.28 \\
    & Selector &                    47.47 $\pm$ 0.14 & 21.02 $\pm$ 0.55 & 73.52 $\pm$ 1.12 & 17.03 $\pm$ 0.24 & 6.34 $\pm$ 0.25 & 30.16 $\pm$ 0.75 & 3.84 $\pm$ 0.39 & 1.01 $\pm$ 0.20 & 8.23 $\pm$ 1.33 \\
    \rowstyle{\color{LightGray}}
    & Best Possible ($\Phi_c$) &                                   64.63 $\pm$ 0.10 & 10.66 $\pm$ 0.06 & 72.34 $\pm$ 0.09 & 64.63 $\pm$ 0.10 & 10.66 $\pm$ 0.06 & 72.34 $\pm$ 0.09 & 64.63 $\pm$ 0.10 & 10.66 $\pm$ 0.06 & 72.34 $\pm$ 0.09\\
    
    \mymidrule
    
    \multirow{5}{*}{CLIP-ViL} & | &                                      47.68 $\pm$ 0.24 & 29.99 $\pm$ 0.13 & 100.00 $\pm$ 0.00 & -153.27 $\pm$ 1.32 & 29.99 $\pm$ 0.13 & 100.00 $\pm$ 0.00 & -2162.82 $\pm$ 12.26 & 29.99 $\pm$ 0.13 & 100.00 $\pm$ 0.00  \\
    & MaxProb &                  54.77 $\pm$ 0.15 & 19.84 $\pm$ 0.38 & 81.98 $\pm$ 0.81 & 21.93 $\pm$ 0.50 & 5.93 $\pm$ 0.24 & 38.47 $\pm$ 1.01 & 2.82 $\pm$ 0.78 & 0.98 $\pm$ 0.24 & 7.27 $\pm$ 2.00 \\
    & Calibration &                        55.00 $\pm$ 0.16 & 18.91 $\pm$ 0.50 & 80.24 $\pm$ 1.09 & 23.16 $\pm$ 0.33 & 5.20 $\pm$ 0.47 & 36.73 $\pm$ 1.79 & 5.29 $\pm$ 0.71 & 0.78 $\pm$ 0.20 & 9.96 $\pm$ 2.35 \\
    & Selector &                   55.47 $\pm$ 0.17 & 18.18 $\pm$ 0.54 & 79.09 $\pm$ 1.07 & 25.93 $\pm$ 0.28 & 5.41 $\pm$ 0.48 & 39.55 $\pm$ 1.96 & 8.00 $\pm$ 0.68 & 0.60 $\pm$ 0.17 & 11.37 $\pm$ 2.11 \\
    \rowstyle{\color{LightGray}}
    & Best Possible ($\Phi_c$) &                                    70.01 $\pm$ 0.13 & 9.86 $\pm$ 0.08 & 77.67 $\pm$ 0.12 & 70.01 $\pm$ 0.13 & 9.86 $\pm$ 0.08 & 77.67 $\pm$ 0.12 & 70.01 $\pm$ 0.13 & 9.86 $\pm$ 0.08 & 77.67 $\pm$ 0.12 \\
    \bottomrule
    \end{tabular}}
\caption{Mean and standard deviation for Effective Reliability $\Phi_c$ over 10 trials from \tabref{tab:effective}. All in \%. See \secref{sec:standardDeviations}.
}
\label{tab:effective-appendix}
\end{table*}

\section{Proof of Lemma 1}
\label{sec:proof}
Lemma 1 states that if a model abstains ``perfectly'', the introduced Effective Reliability score is equal to the VQA Accuracy. 
In this section, we provide a proof of Lemma 1 in the main paper, which we repeat here for ease of understanding the proof:

\myparagraph{Lemma 1.} \emph{The Effective Reliability score is equal to the VQA Accuracy ($\Phi_c(x)= Acc(x)$) if a model abstains ($g(x)=0$) \emph{iff} it is incorrect ($Acc(x)=0$).}

Distilling this to the mathematical notation:
\begin{equation}
     (g(x)=0 \leftrightarrow Acc(x)=0) \longrightarrow \Phi_c(x)= Acc(x)
\end{equation}

Extending Eq. 6 to both cases, $Acc(x)=0$ and $Acc(x)>0$ (note, that Acc cannot be smaller than 0):

\begin{equation}\label{eq:effectiveness_extended}
        \Phi_c(x) = 
    \begin{cases}
    Acc(x)& \text{if}\  g(x) = 1\  \text{and}\  Acc(x) > 0,\\
    -c & \text{if}\ g(x) = 1\  \text{and}\  Acc(x) = 0,\\
    0 & \text{if}\ g(x) = 0\ \text{and}\  Acc(x) > 0,\\
    0 & \text{if}\ g(x) = 0\ \text{and}\  Acc(x) = 0.
    \end{cases}
\end{equation}

To prove Lemma 1, we must show that the condition $(g(x)=0 \leftrightarrow Acc(x)=0)$ implies $\Phi_c(x)= Acc(x)$. The condition $(g(x)=0 \leftrightarrow Acc(x)=0)$ simplifies \eqnref{eq:effectiveness_extended} as the second and third line contradict the condition:
\begin{equation}\label{eq:effectiveness_extended_condidtional}
        \Phi_c(x) = 
    \begin{cases}
    Acc(x)& \text{if}\  g(x) = 1\  \text{and}\  Acc(x) > 0,\\
    0 & \text{if}\ g(x) = 0\ \text{and}\  Acc(x) = 0.\\
    \end{cases}
\end{equation}
As the $Acc(x) = 0$, the second line can be re-written as:
\begin{equation}\label{eq:effectiveness_extended_condidtional2}
        \Phi_c(x) = 
    \begin{cases}
    Acc(x)& \text{if}\  g(x) = 1\  \text{and}\  Acc(x) > 0,\\
    Acc(x) & \text{if}\ g(x) = 0\ \text{and}\  Acc(x) = 0.\\
    \end{cases}
\end{equation}
Now, in both cases $\Phi_c(x)= Acc(x)$ \hspace{3.5cm} $\qed$

\section{Relation to Conformal Prediction}
\label{sec:conformalprediction} Conformal prediction aims to predict a set of outputs, with a guarantee that the set contains the correct output with a specified probability~\cite{vovk2005algorithmic,shafer2008tutorial}.
In VQA, the criterion of a set containing the ``correct output'' is harder to define. For example, two distinct answers might be both be true (\textit{``yellow'',``brown''}) for \textit{``What color are the bananas?''}, but others sets might be contradictory (\textit{``yes'',``no''}). Further research might focus on how to best convey answer sets to users in VQA and how semantic similarity of answers should be modeled, or on the design of better criteria to determine a set-based risk.
More generally, the field of risk control, which does not require variable-size output sets, provides theoretical guarantees that a given error measure is below a tolerance level with some specified probability~\cite{angelopoulos2021gentle,ji2021test}.
\cite{angelopoulos2021learn} describes how to choose a prediction threshold to satisfy a guarantee on error bound.
\cite{ji2021test} relates these guarantees to test sample accuracy based on training sample density.
We view these probabilistic guarantees on error bounds as complementary to our framework, with opportunities for future work to incorporate them both.

\end{document}